\documentclass[lettersize,journal]{IEEEtran}
\usepackage{amsmath,amsfonts}
\usepackage{algorithmic}
\usepackage{algorithm}
\usepackage{array}
\usepackage[caption=false,font=normalsize,labelfont=sf,textfont=sf]{subfig}
\usepackage{textcomp}
\usepackage{stfloats}
\usepackage{url}
\usepackage{verbatim}
\usepackage{graphicx}
\usepackage{cite}
\usepackage[font=small,labelfont=bf]{caption}
\usepackage{orcidlink}
\hypersetup{
    colorlinks=false,
    pdfborder={0 0 0},
}

\ifCLASSOPTIONcompsoc
  \usepackage[nocompress]{cite}
\else
  \usepackage{cite}
\fi
\ifCLASSINFOpdf
\else
\fi
\usepackage{array}
\usepackage{url}

\usepackage{times}
\usepackage{latexsym}
\usepackage{caption}
\usepackage{amsfonts}
\usepackage{amsmath}
\usepackage[utf8]{inputenc}

\usepackage{graphicx, subfig}
\usepackage{booktabs}
\usepackage{amssymb}
\usepackage{eufrak}
\usepackage{mathalfa}
\usepackage{makecell}
\usepackage{lipsum}
\usepackage{marvosym}
\usepackage{algorithmic,algorithm}
\usepackage{multirow}

\usepackage{multicol} 
\usepackage{xcolor}
\usepackage{tabularx}
\usepackage{amsthm}
\usepackage{bm}
\usepackage{academicons}
\usepackage{microtype}

\begin{document}
%
\title{Improving Pre-trained Language Model Fine-tuning with Noise Stability Regularization}
%
%
%
%

\author{Hang~Hua\ \orcidlink{0000-0002-5441-5776},
        Xingjian~Li\ \orcidlink{0000-0001-8073-7552}~\IEEEmembership{}
        Dejing~Dou\  \orcidlink{0000-0001-7561-1672},~\IEEEmembership{Senior Member,~IEEE} \\
        Cheng-Zhong~Xu\ \orcidlink{0000-0001-9480-0356},~\IEEEmembership{Fellow,~IEEE,}
        and~Jiebo~Luo\ \orcidlink{0000-0002-4516-9729},~\IEEEmembership{Fellow,~IEEE}




\thanks{H. Hua and J. Luo are with University of Rochester, Rochester, NY\ 14627\ USA (
E-mail: \{hhua2,\ jluo\}@cs.rochester.edu).}
\thanks{X. Li is with Carnegie Mellon University, Pittsburgh, PA 15213 USA (
E-mail: xingjia2@andrew.cmu.com).}
\thanks{D. Dou is with BCG in Greater China, Beijing, 100027, China. (
E-mail: dou@cs.uoregon.edu).}
\thanks{C.Z Xu is with the State Key Lab of IOTSC, Faculty of Science and Technology, University of Macau, Macau SAR 999078, China (
E-mail: czxu@um.edu.mo).}
\thanks{H. Hua and X. Li contributed equally. Correspondences to J. Luo.
}}

%
%

\markboth{Journal of \LaTeX\ Class Files,~Vol.~14, No.~8, August~2021}%
{Shell \MakeLowercase{\textit{et al.}}: A Sample Article Using IEEEtran.cls for IEEE Journals}

\maketitle
\begin{abstract}
\NoHyper
The advent of large-scale pre-trained language models has contributed greatly to the progress in natural language processing. Despite its recent success and wide adoption, fine-tuning a pre-trained language model often suffers from overfitting, which leads to poor generalizability due to the extremely high complexity of the model and the limited training samples from downstream tasks. To address this problem, we propose a novel and effective fine-tuning framework, named \textbf{L}ayerwise \textbf{N}oise \textbf{S}tability \textbf{R}egularization (\textbf{LNSR}). Specifically, our method perturbs the input of neural networks with the standard Gaussian or In-manifold noise in the representation space and regularizes each layer's output of the language model. We provide theoretical and experimental analyses to prove the effectiveness of our method. The empirical results show that our proposed method outperformes several state-of-the-art algorithms such as $\text{L}^2$-SP~\cite{Li2018ExplicitIB}, Mixout~\cite{Lee2020MixoutER}, FreeLB \cite{Zhu2020FreeLBEA} and SMART~\cite{Jiang2020SMARTRA}, etc. In addition to evaluating the proposed method on relatively simple text classification tasks, similar to the prior works, we further evaluate the effectiveness of our method on more challenging question-answering tasks. These tasks present a higher level of difficulty, and they provide a larger amount of training examples for tuning a well-generalized model. Furthermore, the empirical results indicate that our proposed method can improve the domain generalization performance of language models on unseen domain data.
\end{abstract}



\NoHyper



\section{Introduction}
\label{sec:introduction}

\IEEEPARstart{L}{arge-scale} pre-trained language models (PLMs) have significantly boosted state-of-the-art performance on Natural Language Processing (NLP) tasks \cite{Guu2020REALMRL, Liu2019FinetuneBF, Wadden2019EntityRA, Zhu2020IncorporatingBI ,clark2020electra,joshi2020spanbert}. In particular, some recently emerged powerful language models \cite{2020t5,lewis2019bart,clark2020electra} with impressing performance on natural language understanding (NLU) tasks in popular NLP benchmarks such as GLUE \cite{Wang2018GLUEAM}, Super GLUE \cite{Wang2019SuperGLUEAS}, LAMA \cite{petroni2019language,petroni2020context}, and variants of these
models have been successfully applied in ever-wide scenarios
\cite{lample2019cross,joshi2020spanbert,hu2022promptcap,lin2023videoxum,zhang2020graph}. 

Fine-tuning is the prevalent paradigm for utilizing large pre-trained language models to perform downstream tasks. By initializing with the pre-trained model, the new task reuses most of the well-learned parameters, thus preserving the intrinsic generalizable knowledge while pursuing adaptation to the desired domain. However, despite the simplicity and ubiquity of fine-tuning in modern NLP, this process is brittle \cite{Devlin2019BERTPO}, i.e., a straightforward fine-tuning process sometimes leads to unstable solutions that generalize poorly to unseen data. Empirical studies have discovered that randomness brought by data order and weight initialization causes unexpected results~\cite{Dodge2020FineTuningPL}. Nevertheless, ad-hoc strategies such as seed selection and early stopping~\cite{Dodge2020FineTuningPL} provide a neither theoretical nor practical guarantee. A systematic solution to this challenge, especially in conditions where labeled examples are insufficient, is needed. 

\begin{figure*}[t]
    \centering
    \includegraphics[width=0.4\linewidth]{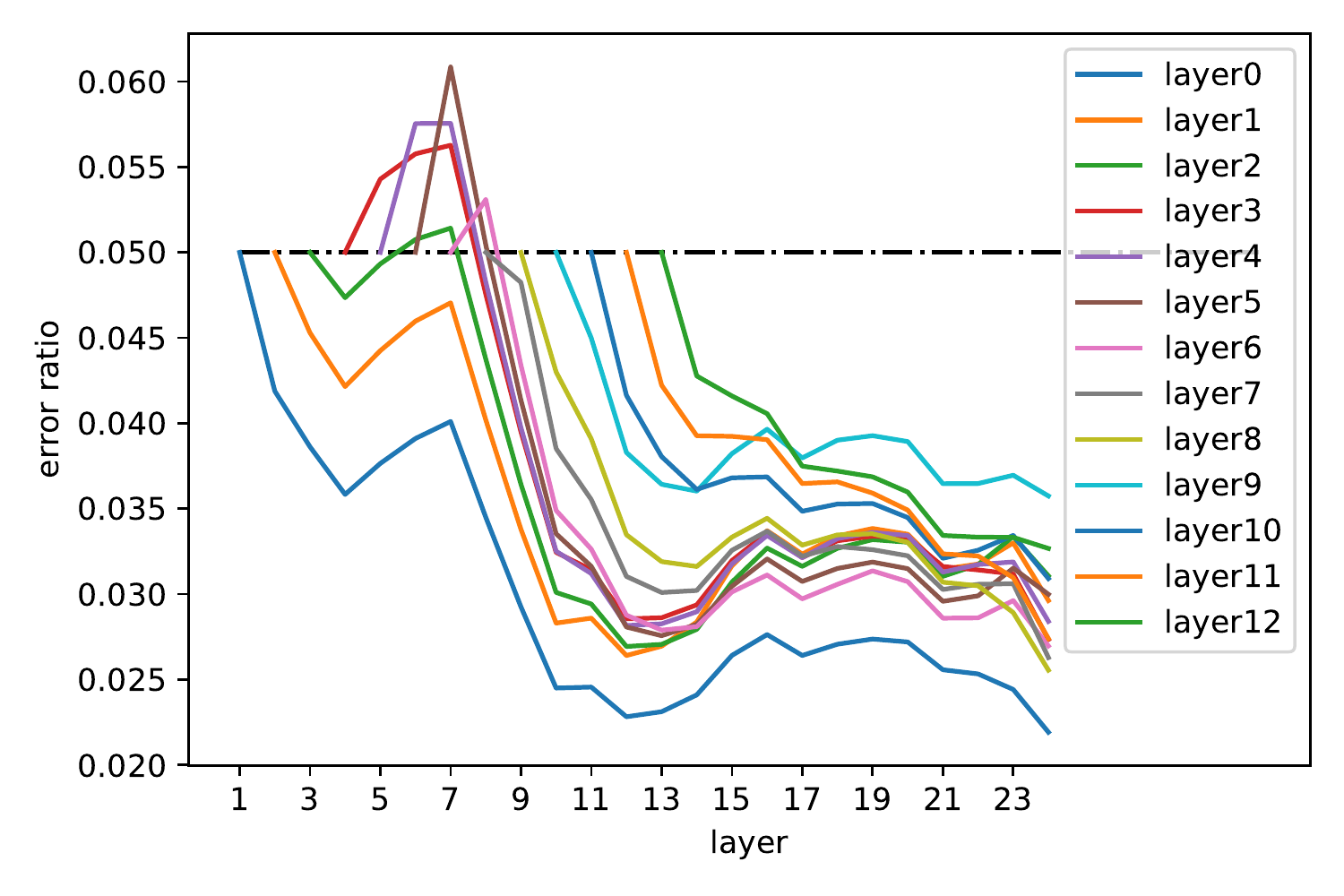}
    \includegraphics[width=0.4\linewidth]{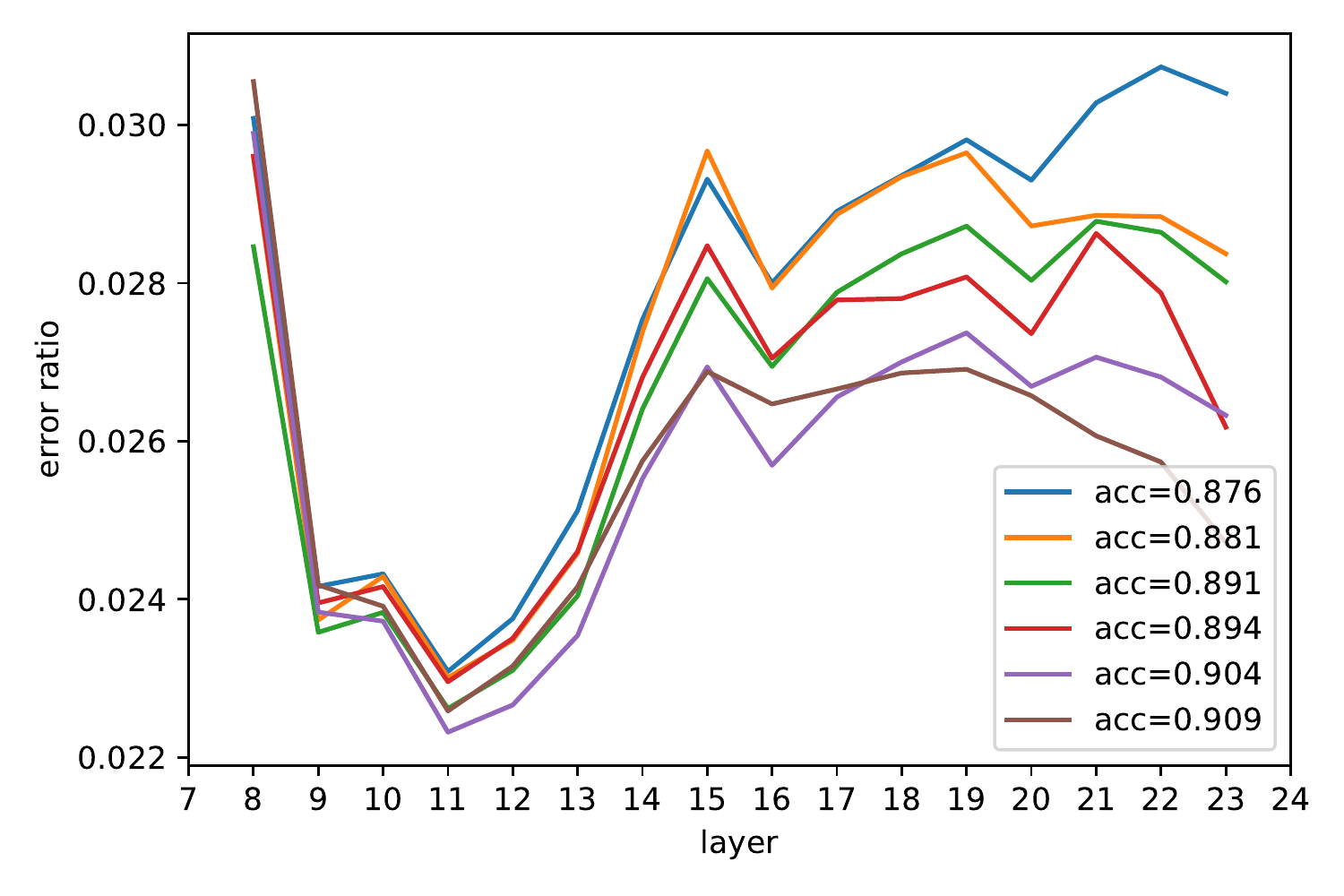}
    \caption{Demonstration of how injected noise attenuates on a RoBERTa-Large model fine-tuned on the MRPC dataset. The error ratio (Y-axis) is defined as the relative output deviation from the original observation on each layer (X-axis). In the \textbf{left} plot, we show behaviors of noise stability by injecting noise at different positions, i.e. each injected position corresponds to a curve. The noise has a random direction, whose magnitude is $5\%$ of that of the original input. We observe that though the propagated perturbation decreases rapidly on lower layers, but becomes volatile on the higher layers (e.g. layers 14-20), indicating the poor robustness and risk of over-fitting on these higher layers. In comparison, we present decent behaviors of noise stability on VGG-19 in Appendix A. We also show in the \textbf{right} plot that, models more robust to input perturbations (noise is injected at layer 1) tend to deliver higher accuracies. Specifically, each curve represents a fine-tuned model, whose accuracy is marked in the legend. The accuracy shows obviously positive correlation with noise stability, especially for higher layers.}
    \label{fig:noise}
\vspace{-3mm}
\end{figure*}

Improving generalization is always one of the fundamental goals of machine learning. Adding noise to the input~\cite{bishop1995training,rifai2011adding} has been proven to have an equivalent effect to training over clean input with an additional regularization term to constrain the solution space.  Recent work~\cite{arora2018stronger,dong2021should} discovers that the generalization capacity of deep neural networks is theoretically linked with the so-called interlayer cushion, characterized by noise sensitivity of the network w.r.t. input. While deep convolutional networks exhibit decent behaviors of noise stability as shown in~\cite{arora2018stronger}, we find that it is not exactly the case for transformer-based language models, e.g., BERT, which has more complex multi-head self-attention architectures. A preliminary experiment is conducted to investigate the sensitivity to the Gaussian noise on transformers. From the results shown in Figure~\ref{fig:noise}, two important observations can be made as follows. 

\begin{itemize}
\item The propagating noise, if injected in lower layers\footnote{In this paper, we use the term \emph{lower layers} to denote layers close to the input and \emph{higher layers} to denote those close to the output.}, can be amplified in some higher layers of BERT.
\item Noise stability of a higher layer has a roughly positive correlation with the generalization performance.
\end{itemize}

Motivated by the above observations, we introduce a new framework of noise stability regularization to improve pre-trained language models' fine-tuning in this work. Specifically, we impose an additional optimization term that forces higher layers of BERT to be resilient to a Gaussian noise injected on lower layers, named Layer-wise Noise Stability Regularization (LNSR) \cite{hua2021noise,li2022method}. The proposed regularization term has a good theoretical property of smoothing the learned function. We further design an advanced implementation of LNSR that generates a random noise with directions restricted by neighborhoods of the input point. The qualitative analysis demonstrates its equivalence to noise stability w.r.t. a Gaussian noise injected in the data manifold, thus referred to as In-manifold Layer-wise Noise Stability Regularization (In-manifold LNSR). 

The main contributions of this paper can be summarized as follows. 
\begin{itemize}
\item Our work is the first step in the investigation of noise stability properties on transformer-based architectures, which are of great interest in natural language processing and computer vision applications. We empirically extend observations about noise stability on fully connected networks and deep convolutional networks to transformers. 

\item We propose two alternative implementations of noise stability regularization. Different from earlier works that directly use perturbed input examples to fit the labels, our method adopts a novel layer-wise regularization that explicitly enforces noise stability of middle layers. Based on this idea, we further present a more effective In-manifold noise stability regularization. Specifically, the sampled noise is restricted in the region formed by interpolations between the input point and its nearest neighborhoods. Under commonly accepted assumptions, the simple method can be regarded as for sampling Gaussian noise in the low-dimensional data manifold. 

\item We provide a detailed theoretical analysis of the noise stability regularization w.r.t. the Gaussian noise, revealing its connection with the Lipschitz continuity and the Tikhonov regularizer. The proposed noise stability regularization is also shown to have a form with better optimization properties than the conventional method that simply trains the model over the perturbed inputs. For the In-manifold noise stability, we provide a qualitative analysis of its relationship with manifold learning. 

\item We conduct extensive experiments on several popular NLP tasks, covering different task types (text classification and question answering) and a wide range of dataset scales (from $\sim 10^2$ to $\sim 10^6$). We compare our approach with state-of-the-art methods aimed at improving fine-tuning pre-trained language models such as $\text{L}^2$-SP~\cite{Li2018ExplicitIB}, Mixout~\cite{Lee2020MixoutER} and SMART~\cite{Jiang2020SMARTRA}. Our approach not only consistently improves the overall performance but also obtains more stable fine-tuning results over multiple random trials. Moreover, our algorithm is also effective in dealing with the risk of domain shift, demonstrated by additional experiments on domain generalization benchmarks. 
\end{itemize}

The remainder of this paper is organized as follows. Section 2 introduces the notations and preliminaries used in this paper. In Section 3, we present the overall framework and implementation details of our method, along with theoretical understanding. In Section 4, we evaluate our method on diverse NLP tasks, including classification, regression, and question answering. In Section 5, we provide further analyses explaining why our approach delivers good performance. In Section 6, we give a brief overview of related studies. Finally, in Section 7, the paper is concluded. 

\begin{table*}[t]
\center
\caption{General Notations}
\small
\begin{tabularx}{0.8\textwidth}{p{0.15\textwidth}X}
\toprule
  \multicolumn{2}{l}{{\underline{Variables:}}} \\
  $d$ & the dimensionality of the input in the original representation space \\
  $(\mathbf{x},\mathbf{y})$   &  an input point $\mathbf{x} \in \mathbb{R}^d$ and its corresponding label $\mathbf{y}$ \\
  $\bm{\varepsilon}$ & a small random noise with the same dimension as the input $\mathbf{x}$ \\
  $\tilde{\mathbf{x}}$   &  the perturbed input that $\tilde{\mathbf{x}}=\mathbf{x}+\bm{\varepsilon}$  \\
  $\bm{\theta}$ & the parameter of a model \\
  $L$ & the total number of layers in a BERT model \\
  $b$ & the index of a layer where the noise is injected on its input \\
  $r$ & the index of a layer where the noise stability is enforced, $1 \leq b \leq r \leq L$ \\
  $k$ & the number of nearest neighbors of an input $\mathbf{x}$ \\
  $N_k(\mathbf{x})$ & the set of the $k$-nearest neighbors of $\mathbf{x}$ \\
  \\
  \multicolumn{2}{l}{{\underline{Functions and Operators:}}} \\
  $F$ & a BERT model parameterized with $\bm{\theta}$ \\
  $f$ & a real-valued function in convenience of the theoretical analysis \\
  $\mathcal{L}$ & the loss function\\
  $\mathcal{R}$ & the regularization term \\
  $\|.\|$ & the $L^2$ norm of a vector, i.e. if $\mathbf{x} \in \mathbb{R}^d$, $\|\mathbf{x}\|=\|\mathbf{x}\|_2=\sqrt{\sum_{i=1}^{d}\mathbf{x}_i^2}$ \\
  $\|.\|_F$ & the Frobenius Norm of a matrix, i.e. if $\mathbf{A} \in \mathbb{R}^{m \times n}$, $\|\mathbf{A}\|_F=\sqrt{\sum_{i=1}^{m}{\sum_{j=1}^{n}\mathbf{A}_{ij}}}$ \\
  $\circ$ & the Hadamard product of two matrices $\mathbf{A}$ and $\mathbf{B}$ (with the same dimension) as $(\mathbf{A} \circ \mathbf{B})_{ij} = \mathbf{A}_{ij}\mathbf{B}_{ij}$
  \\
  \multicolumn{2}{l}{{\underline{Constants:}}} \\
    $\mathbf{I}$ & the identity matrix with ones on the main diagonal and zeros elsewhere \\
    $\mathbf{1}$ & the all-ones matrix where every element is equal to one \\
  \bottomrule
\end{tabularx}
\label{table:notations}
\end{table*}

\section{Notations and Preliminaries}
In this section, we will first introduce the notations frequently used in this paper for clear representations. Then we briefly present some preliminary knowledge closely connected with our proposed algorithm and theoretical analysis.  
\subsection{Notations}
Throughout, we will frequently use the set of notations and terminology listed in Table~\ref{table:notations}. 

\subsection{Preliminaries}
\subsubsection{Properties of the Gaussian Distribution}
This part briefly introduces properties of the Gaussian distribution used in the paper. The Gaussian distribution is often referred to as $\mathcal{N}(\mu, \sigma^2)$, where the parameter $\mu$ is the expectation of the distribution, while $\sigma$ is the standard deviation. The general Gaussian distribution $\mathcal{N}(\mu, \sigma^2)$ can be described by its probability density function 
\begin{equation}
p(\varepsilon) = \frac{1}{\sqrt{2\pi} \sigma}e^{-\frac{(\varepsilon-\mu)^2}{2\sigma^2}}. 
\end{equation}

For random vectors, the univariate Gaussian distribution can be generalized to higher dimensions, described by the multivariate Gaussian distribution $\bm{\varepsilon} \sim \mathcal{N}(\bm{\mu},\bm{\Sigma})$, where $\bm{\mu}$ is the mean vector and $\bm{\Sigma}$ is the covariance matrix. In this work, we shall focus on the standard case that $\bm{\mu}$ is a zero vector and $\bm{\Sigma}$ is a diagonal matrix with all diagonal elements being $\sigma^2$, i.e. $\bm{\Sigma}=\sigma^2 \mathbf{I}$.

Note that the $d$-dimensional standard multivariate Gaussian distribution $\bm{\varepsilon} \sim \mathcal{N}(\bm{0},\sigma^2 \mathbf{I})$ has the following properties:
\begin{equation}
    \forall i,j \in [1,d], \underset{\bm{\varepsilon}}{\mathbb{E}} \{ \bm{\varepsilon}_i\bm{\varepsilon}_j\}=\sigma^2 \delta_{ij},
\end{equation}
and
\begin{equation}
    \forall i,j,k \in [1,d], \underset{\bm{\varepsilon}}{\mathbb{E}} \{ \bm{\varepsilon}_i\bm{\varepsilon}_j\bm{\varepsilon}_k\}=0,
\end{equation}
which will be frequently used in our theoretical analysis. 

\subsubsection{Lipschitz Continuity}
\label{sec:pre_lip}

In mathematical analysis, Lipschitz continuity is used to quantify the degree of smoothness of a function. Intuitively, it describes how fast can the output change as the input changes. It's of great interest to the machine learning community because a smooth function is often regarded as generalizing well to unseen data. A formal definition is presented as follows. 

\noindent \textbf{Definition 1.} Given two metric spaces $(X,d_X)$ and $(Y,d_Y)$, a function $f:X \to Y$ is called Lipschitz continuous if there exists a real constant $K \ge 0$ such that, for all $x_1, x_2 \in X$,
\begin{equation}
    d_Y(f(x_1),f(x_2)) \le Kd_X(x_1,x_2).
\end{equation}
The smallest $K$ satisfying the previous inequality is called the Lipschitz constant of $f$, denoted $\mathrm{Lip}(f)$. In this work, we consider only the common case of the Hilbert space on $\mathbb{R}^n$ equipped with the distance metric $d(a, b) = \|a - b\|$. Under an assumption that $f$ is locally Lipschitz, \cite{federer1996geometric} proposes an approach to estimate the Lipschitz constant by the following theorem. 

\noindent \textbf{Theorem 1} (from \cite{federer1996geometric}). \emph{If $f:\mathbb{R}^d \to \mathbb{R}^m$ is a locally Lipschitz continuous function, then $f$ is differentiable almost everywhere. Moreover, if $f$ is Lipschitz continuous, then}
\begin{equation}
\label{eq:lip_jacobian}
    \mathrm{Lip}(f) = \underset{\mathbf{x} \in \mathbb{R}^d}{\mathrm{sup}}\|J_f(\mathbf{x})\|_2,
\end{equation}
where $J_f(\mathbf{x})$ denotes the differential operator, also called the \emph{Jacobian}, of $f$ at $\mathbf{x}$. Generally $J_f(\mathbf{x})$ is a matrix that $J_f(\mathbf{x}) \in \mathbb{R}^{m \times d}$ and $\|.\|_2$ is  the spectral norm of a matrix $\mathbf{A}$ as
\begin{equation}
    \|\mathbf{A}\|_2 = \underset{\mathbf{x} \in \mathbb{R}^d}{\mathrm{sup}}\frac{\|\mathbf{A}\mathbf{x}\|_2}{\|\mathbf{x}\|_2}.
\end{equation}

\subsubsection{Training with Noise}
Training with noise has been well-studied in previous works. Earlier work~\cite{sietsma1991creating} proposes to add a random vector onto the input before being fed to the neural network, leading to improved generalization performance. \cite{bishop1995training} theoretically proves that training with Gaussian noise is equivalent to Tikhonov regularization, which aims at making the loss surface flatter at the input, through involving derivatives of the objective function w.r.t. different orders. When considering a scalar input variable $x$ and output variable $y$, the Tikhonov regularizer~\cite{tikhonov1977solutions} takes a general form  
\begin{equation}
    \mathcal{R}_{\mathrm{Tik}}(\theta) = \sum_{r} \int h_r(x) (\frac{\partial^r f}{\partial x^r})^2 dx.
\end{equation}

Specifically, if sampling the noise from a distribution that has zero means and is independent between different inputs, training over perturbed inputs~\cite{bishop1995training} equals to training over clean inputs with an extra regularization term 
\begin{equation}
    \mathcal{R}(\theta) \approx \underset{(\mathbf{x},\mathbf{y})}{\mathbb{E}} \{ (f(\mathbf{x};\bm{\theta})-\mathbf{y}) \mathrm{Tr}(H(\mathbf{x})) + \|J(\mathbf{x})\|^2 \}.
\end{equation}

A drawback of simply adding noise to inputs~\cite{bishop1995training} is that the Hessian term is not guaranteed to be positive. Such an unconstrained term used as a regularizer may lead to a negative Hessian trace on a very large scale. ~\cite{rifai2011adding} proposes an improved regularized objective that adds noise to inputs of the Jacobian function $\|J(\tilde{\mathbf{x}};\bm{\theta})\|^2$. By ignoring the higher order of derivative, the perturbed Jacobian induces an approximated regularizer as
\begin{equation}
    \mathcal{R}(\theta) \approx \underset{\mathbf{x}}{\mathbb{E}} \{\|J(\mathbf{x})\|^2 + 2\sigma^2\|H(\mathbf{x})\|^2_F \},
\end{equation}
where $\sigma$ is the variance of the noise. Though this new noise term is proved to avoid the undesired effect of \cite{bishop1995training}, involving perturbed Jacobian in the optimization objective requires so-called \emph{double back-propagation}. This encounters considerable computational inefficiency for modern DNNs such as BERT. 

\subsubsection{Dimensionality Reduction and Manifold Learning}
Dimensionality reduction plays a crucial role in enhancing human understanding and processing of real-world data, particularly when confronted with high-dimensional and intricate structures. However, traditional approaches like Principal Components Analysis (PCA) are limited to linear structures. In contrast, manifold learning has emerged as a powerful framework for non-linear dimensionality reduction~\cite{roweis2000nonlinear,lin2008riemannian}, operating under the assumption that the target data resides in a locally Euclidean topological space~\cite{huo2007survey,cayton2005algorithms}. Manifold learning techniques enable the visualization of high-dimensional data~\cite{van2008visualizing} in a lower-dimensional space, such as a two or three-dimensional map, facilitating intuitive interpretation and analysis.

The In-manifold LNSR algorithm draws inspiration from the aforementioned assumption that data resides in a low-dimensional manifold, which is locally homeomorphic to the original representation space. Based on this principle, directly sampling noise in the original high-dimensional space may prove inefficient since many features are irrelevant to the target data. To address this issue, we adopt an efficient perturbation strategy through In-manifold noise sampling, enabling effective modifications to the input. It is important to note that our approach leverages the idea of the manifold assumption but does not aim to implement dimensionality reduction algorithms.

\section{Methodology}
In this section, we systematically introduce our algorithm composed of the following parts. We first present the general framework of our proposed Layer-wise Noise Stability Regularization (LNSR) for BERT in subsection 3.1. Next, we describe two alternative methods of noise generalization in 3.2, which are the \emph{Standard} LNSR that directly injects Gaussian noise, and \emph{In-manifold} LNSR that adds random noise constrained on the subspace formed by the input's nearest neighbors. Then, we provide a theoretical analysis for the two specific choices of noise generation methods. In 3.3, we prove that the  \emph{Standard} LNSR has good properties related with the Lipschitz continuity and Tikhonov regularization. In 3.4, we demonstrate that the \emph{In-manifold} LNSR is equivalent to the \emph{Standard} LNSR imposed on the data manifold under certain assumptions. 

\begin{algorithm}[h]
\caption{Fine-tuning with LNSR Regularization} 
\begin{algorithmic}[1]
\REQUIRE Training set $\mathcal{D}$, learning rate $\tau$,  number of training iterations $N$, number of layers $L$, neural network $f$ and its corresponding parameter $\bm{\theta}$, the layer index $b$ where noise is injected, and 
regularization weights $\{\lambda^{b,r}\}_{r=b}^{L}$.
\STATE Initialize $\bm{\theta}$ with $\bm{\theta}_0$ learned from the pre-trained task
\FOR{iteration=$1,2,...,N$ } 
    \STATE sample a batch of data $B \sim \mathcal{D}$
    \STATE $\mathcal{R}\gets 0$
    \FOR{each $\mathbf{x} \in B$}
        \IF{\emph{Standard} LNSR}
        \STATE $\bm{\varepsilon} \sim \mathcal{N}(0, \sigma^2 \mathbf{I})$
        \ELSIF{\emph{In-manifold} LNSR}
        \STATE get $k$-nearest neighborhoods $N_k(\mathbf{x})=\{\mathbf{x}^{(j)}\}$
        \STATE get differences $\{\mathbf{d}^{(j)}|\mathbf{d}^{(j)}=\mathbf{x}^{(j)}-\mathbf{x}\}$
        \STATE orthogonalize $\{\mathbf{d}^{(j)}\}$ and get $\{\check{\mathbf{d}}^{(j)}\}$
        \STATE perform $k$ iid sampling of $\{\varepsilon^{(j)}| \varepsilon^{(j)} \sim \mathcal{N}(0, \sigma^2)\}$
        \STATE $\bm{\varepsilon}=\sum_{j=1}^{k}\varepsilon^{(j)} \check{\mathbf{d}}^{(j)}$
        \ENDIF
        \STATE $\tilde{\mathbf{x}}\gets \mathbf{x}+\bm{\varepsilon}$
        \STATE feed $\mathbf{x}$ and $\tilde{\mathbf{x}}$ into the network $f$
            \FOR{$r=b,b+1,...,L$}
            \STATE $\mathcal{R}\gets \mathcal{R}+\lambda^{b,r}||f^{b,r}(\mathbf{x})-f^{b,r}(\tilde{\mathbf{x}})||^2$
            \ENDFOR
        \ENDFOR
    \STATE $g \gets \frac{1}{|B|}\sum_{(\mathbf{x},y)}
    \nabla[\mathcal{L}(f(\mathbf{x};\bm{\theta}),y)+\mathcal{R}] $
    \STATE$ \bm{\theta}\gets \bm{\theta}-\tau g $
\ENDFOR
\ENSURE $\bm{\theta}$
\end{algorithmic}
\label{algo:frame}
\end{algorithm}
\vspace{-6mm}

\subsection{The General Framework}
Given a pre-trained model as initialization, fine-tuning BERT is a general task of supervised learning that aims at minimizing an expected error $\mathcal{L}$ with respect to the model's parameter $\bm{\theta}$ over the data distribution. Considering the high memorization capacity of deep neural networks, a regularization term $\mathcal{R}$ responsible to control the model's complexity is often employed to improve the generalized performance of the model. Therefore, a general form of the optimization objective can be represented as
\begin{equation}
    \bm{\theta}^*=\mathop{\arg\min}_{\bm{\bm{\theta}}} \underset{(\mathbf{x},y)}{\mathbb{E}}[ \mathcal{L}(f(\mathbf{x};\bm{\theta}),y)]+ \mathcal{R}(\bm{\theta}),
\end{equation}
where we omit the notation of the data distribution without ambiguity. A most common choice for the regularization $\mathcal{R}$ is the $L^2$ normalization of the parameter $\bm{\theta}$, that is $\|\bm{\theta}\|^2$, also called the weight decay, particularly for deep neural networks. Despite its ubiquity, there's no theoretical evidence for the effectiveness of such a simple data-independent regularization in DNNs~\cite{van2017l2,zhang2018three}.

We are interested in the behavior of noise stability, which serves as a data-dependent regularizer. Specifically, given an input point $\mathbf{x}$, we generate a perturbed input $\tilde{\mathbf{x}}$ by adding a random noise $\bm{\varepsilon}$ with a small magnitude to $\mathbf{x}$. Noise stability characterizes to what degree the output w.r.t. $\tilde{\mathbf{x}}$ deviates from that w.r.t. the clean input $\mathbf{x}$. For the multi-layer transformer architecture, we adopt a layer-wise regularization to enforce noise stability at each layer. Formally, we define the input of layer $b$ as $\mathbf{x}^b$ and the network between the $b$-th and $r$-th layer as $f^{b,r}$, which is parameterized by $\bm{\theta}^{b,r}$. Note that when $b=r$, $f^{b,r}$ represents a single layer. If we inject the noise at a fixed layer $b$ and regularize each higher layer $r \geq b$, the noise stability term can be represented as 
\begin{equation}
\label{eq:LNSR}
    \mathcal{R}(\bm{\theta}) = \underset{\mathbf{x},\bm{\varepsilon}}{\mathbb{E}} \sum_{r=b}^L \lambda^{b,r} ||f^{b,r}(\mathbf{x}^b+\bm{\varepsilon};\bm{\theta}^{b,r})-f^{b,r}(\mathbf{x}^b;\bm{\theta}^{b,r})||^2,
\end{equation}
where $\lambda^{b,r}$ is the coefficient to control the weight of regularizing $f^{b,r}$. Given Eq.~\ref{eq:LNSR} as the general form of noise stability used in multi-layer transformers, a subsequent question is how to generate the noise $\bm{\varepsilon}$ for a specific input $\mathbf{x}^b$.

\begin{figure}[t]
    \centering
    \includegraphics[width=0.49\linewidth]{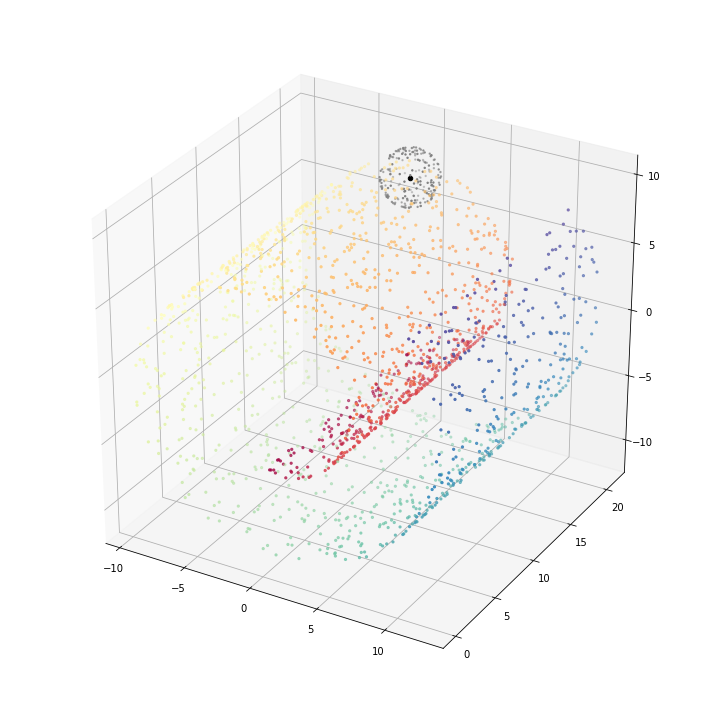}
    \includegraphics[width=0.49\linewidth]{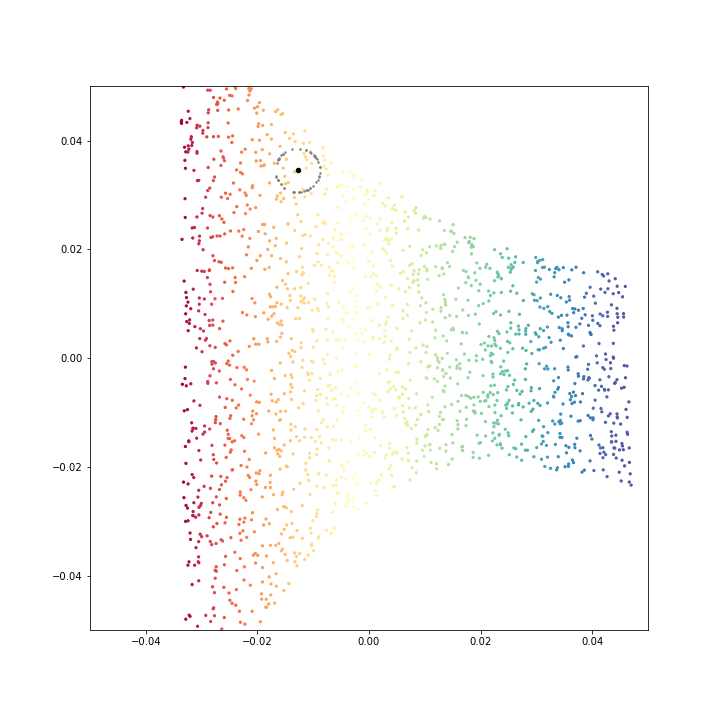}
    \caption{Illustration of noise stability performed in the original $\mathbb{R}^d$ space (\textbf{left}) and the data manifold (\textbf{right}). The black and grey dots refer to the clean and perturbed data respectively. In the original space, the perturbed data, lying on a unit sphere around the clean input, would likely to be out-of-domain. In contrast, adding in-manifold noise to the clean data is probably to yield a perturbed point with high data density. In this sense, in-manifold perturbation more accurately simulates data shifts between training and test data and, as a result, delivers better generalization performance on unseen data. }
    \label{fig:manifold}
\end{figure}

\subsection{Methods for Noise Generation}
In this work, we introduce two progressive methods for noise generation. The first sample noise from the multivariate Gaussian distribution for the noise stability regularization is called \emph{Standard} LNSR. We also propose an improved method called \emph{In-manifold} LNSR, which samples random noise on the subspace formed by the input's $k$-nearest neighbors. 

\subsubsection{Standard LNSR}
For the Standard LNSR, $\bm{\varepsilon}$ is randomly sampled from the standard multivariate Gaussian distribution as $\bm{\varepsilon} \sim \mathcal{N}(\bm{0},\sigma^2 \mathbf{I})$. That is, each element $\bm{\varepsilon}_i$ is independently sampled from the zero-mean Gaussian distribution as $\bm{\varepsilon}_i \sim \mathcal{N}(0, \sigma^2)$. 

\subsubsection{In-manifold LNSR}
In this section, we introduce the implementation of In-manifold LNSR in detail with intuitive explanations. More formal discussions are deferred to 3.4. A widely accepted assumption for a smooth manifold is that, an input point $\mathbf{x}$ and its $k$ neighborhoods $N_k(\mathbf{x})$ form a subspace that is approximated linearly. With $\{\mathbf{x}^{(j)}|\mathbf{x}^{(j)} \in N_k(\mathbf{x}), j=1,2,...,k\}$ denoting these neighborhoods, we get the set of neighbored differences as $\{\mathbf{d}^{(j)}=\mathbf{x}^{(j)}-\mathbf{x}|\mathbf{x}^{(j)} \in N_k(\mathbf{x}), j=1,2,...,k\}$. Under the manifold assumption, vectors in $\{\mathbf{d}^{(j)}\}$ lie on the same linear subspace around $\mathbf{x}$ as the origin. Imagining concatenating all these row vectors $\{\mathbf{d}^{(j)}\}$ to form a matrix, we perform the Gram-Schmidt orthogonalization on the matrix and obtain the transformed set $\{\check{\mathbf{d}}^{(j)}\}$ which are all orthogonal to each other. It's obvious that $\{\check{\mathbf{d}}^{(j)}\}$ lie on the same subspace as $\{\mathbf{d}^{(j)}\}$. To generate In-manifold random noise, we independently sample $k$ random variables from the univariate Gaussian distribution as $\varepsilon^{(j)} \sim \mathcal{N}(0, \sigma^2)$. Then we generate the In-manifold noise $\bm{\varepsilon}$ by a random interpolation in the region formed by $\{\check{\mathbf{d}}^{(j)}\}$ as $\bm{\varepsilon}=\sum_{j=1}^{k}\varepsilon^{(j)} \check{\mathbf{d}}^{(j)}$. Figure~\ref{fig:manifold} demonstrates the difference between sampling  standard Gaussian noise and sampling In-manifold noise. 

\subsection{Analysis of the Standard LNSR}
Here we present mathematical analyses that highlight the properties of our proposed approach. Our primary goal is to examine the relationships between LNSR and classical techniques used for quantifying and controlling model complexity.  The reduction of model complexity is widely acknowledged as a crucial principle in mitigating overfitting in machine learning. However, due to the intricate nature of Deep Neural Networks (DNNs), which encompass a massive number of parameters, traditional metrics for quantification often become computationally challenging. Based on the analyses in this part, LNSR is shown to have implicit connections with Lipschitz continuity and Tikhonov regularization.

Without loss of generality, the analysis is based on a general real-valued function $f: \mathbb{R}^d \to \mathbb{R}$. For simplicity, we ignore the notation of the parameter $\bm{\theta}$ in the following analysis. $\bm{\varepsilon}$ is the standard Gaussian noise as defined in 3.2.1. Given that $\bm{\varepsilon}$ has a small magnitude (i.e. its variance $\sigma^2$), we adopt the second-order Taylor approximation to present $f(\mathbf{x}+\bm{\varepsilon})$  as:
\begin{equation}
    f(\mathbf{x} + \bm{\varepsilon}) = 
    f(\mathbf{x}) + J(\mathbf{x}) \bm{\varepsilon}  + 
    \frac{1}{2} \bm{\varepsilon}^T H(\mathbf{x}) \bm{\varepsilon} + O(\bm{\varepsilon}^3),
\label{eq:taylor}
\end{equation}
where $J(\mathbf{x})$ and $H(\mathbf{x})$ are the Jacobian and Hessian of $f$ w.r.t $\mathbf{x}$, respectively. 

\subsubsection{Connection with Lipschitz Continuity}
By ignoring the second and higher order terms in the Taylor expansion, it is easy to derive that the noise stability regularization equals to the $L^2$ norm of the Jacobian $J_f$ 
\begin{equation}
\label{eq:LNSR_lip}
\underset{\mathbf{x},\bm{\varepsilon}}{\mathbb{E}} \|f(\mathbf{x} + \bm{\varepsilon}) - f(\mathbf{x})\|^2 = \underset{\mathbf{x},\bm{\varepsilon}}{\mathbb{E}} \|J(\mathbf{x}) \bm{\varepsilon}\|^2 = \sigma^2 \underset{\mathbf{x},\bm{\varepsilon}}{\mathbb{E}} \frac{\|J(\mathbf{x}) \bm{\varepsilon}\|^2}{\|\bm{\varepsilon}\|^2}.
\end{equation}

As indicated in \ref{sec:pre_lip}, the Lipschitz constant of $f$ is bounded by the supremum of spectral norm of the Jacobian $J_f$. Therefore, injecting input noise and explicitly minimizing the output discrepancy (between the clean and perturbed output) has the same form as Eq.~\ref{eq:lip_jacobian} inside the operation of calculating the supremum. 

Note that the objective of minimizing Eq.~\ref{eq:LNSR_lip} (w.r.t. a random noise) is \emph{likely} to lower the bound in Eq.~\ref{eq:lip_jacobian} but is of course not guaranteed to do so. In fact, $\bm{\varepsilon}$ with a direction that maximizes $\|J(\mathbf{x})\|_2$, or $\|f(\mathbf{x}+\bm{\varepsilon})-f(\mathbf{x})\|_2$, is so-called an \emph{adversarial perturbation} and such a perturbed input $\tilde{\mathbf{x}} = \mathbf{x}+\bm{\varepsilon}$ is called an \emph{adversarial example}~\cite{szegedy2014intriguing}. Though the adversarial example induces a more accurate indicator of the Lipschitz constant, empirical studies show that training with such extreme inputs, aiming at promoting the robustness of adversarial examples, significantly harms the performance on clean inputs~\cite{madry2018towards}. An intuitive explanation is that the adversarial example focuses on a rare perturbation direction (barely seen in real data) that is the most difficult for the model to be robust. In contrast, our noise stability regularization cares about  noise with a uniform direction, which is more diverse and more possible to exist in real data, though not providing the tight Lipschitz continuity bound. 

\subsubsection{Connection with Tikhonov Regularization}
Here we provide a further analysis considering the first and second-order terms in Eq.~\ref{eq:taylor}. Let $\mathrm{Tr}(.)$ be the trace of a matrix. We first present our main claim as follows. 

\hspace*{\fill}

\noindent \textbf{Claim 1.} \emph{If a real-valued function $f : \mathbb{R}^d \to \mathbb{R}$ is twice differentiable with respect to its input $\mathbf{x} \in \mathbb{R}^d$, and $\bm{\varepsilon} \in \mathbb{R}^d$ conforms to the standard multivariate Gaussian distribution $\bm{\varepsilon} \sim \mathcal{N}(0,\sigma^2 \mathbf{I})$. Then, omitting terms of higher order than the 2nd degree, the noise stability regularization $\mathcal{R}$ for a given $\mathbf{x}$ can be represented as}
\begin{equation}
\begin{aligned}
    \mathcal{R} &= \underset{\mathbf{x},\bm{\varepsilon}}{\mathbb{E}}\|f(\mathbf{x}+\bm{\varepsilon})-f(\mathbf{x})\|^2 \\
    &\approx \frac{\sigma^2}{4} \underset{\mathbf{x}}{\mathbb{E}} \{4\|J(\mathbf{x})\|^2 + \|\mathrm{Tr}(H(\mathbf{x}))\|^2 + \|(\mathbf{1}-\mathbf{I}) \circ H(\mathbf{x})\|_F^2 \}.
\end{aligned}
\label{eq:claim}
\end{equation}

\hspace*{\fill}

\begin{proof}
In the following derivations, we simply use $J$ to denote $J(\mathbf{x})$ without ambiguity, and likewise for $H(\mathbf{x})$. $J$ and $H$ with subscripts refer to the partial derivatives, i.e. $J_i = \frac{\partial f(\mathbf{x})}{\partial \mathbf{x}_i}$ and $H_{ij}=\frac{\partial^2 f(\mathbf{x})}{\partial \mathbf{x}_i \partial \mathbf{x}_j}$. 

We begin by substituting the second-order Taylor formula into our definition of the noise stability regularization 
\begin{equation} 
\begin{aligned}
    \mathcal{R} &=  \underset{\mathbf{x},\bm{\varepsilon}}{\mathbb{E}}\{ \|J\bm{\varepsilon} +
    \frac{1}{2} \bm{\varepsilon}^T H \bm{\varepsilon} \|^2 \}.
\label{eq:main}
\end{aligned}
\end{equation}

Using $\Omega_J = J\bm{\varepsilon}$ and $\Omega_H = \frac{1}{2} \bm{\varepsilon}^T H \bm{\varepsilon}$, Eq.~\ref{eq:main} can be notated as 
\begin{equation}
\label{eq:main2}
    \mathcal{R} = \underset{\mathbf{x},\bm{\varepsilon}}{\mathbb{E}} \{\Omega_J^2+\Omega_H^2+2 \Omega_J \cdot \Omega_H\}.
\end{equation}

Since expectation is a linear operator, we reformulate Eq.~\ref{eq:main2} into expectations of three parts as $\mathcal{R}_J=\underset{\mathbf{x},\bm{\varepsilon}}{\mathbb{E}}\{\Omega_J^2\}$, $\mathcal{R}_H=\underset{\mathbf{x},\bm{\varepsilon}}{\mathbb{E}}\{\Omega_H^2\}$ and  $\mathcal{R}_{JH}=2\underset{\mathbf{x},\bm{\varepsilon}}{\mathbb{E}}\{\Omega_J \cdot \Omega_H\}$. 

For the first part $\mathcal{R}_J$, we obtain
\begin{equation}
\begin{aligned}
    \mathcal{R}_J &= \underset{\mathbf{x},\bm{\varepsilon}}{\mathbb{E}} \|J \bm{\varepsilon}\|^2  \\
     &= \underset{\mathbf{x}}{\mathbb{E}} \underset{\bm{\varepsilon}}{\mathbb{E}} \{\sum_{i} (\bm{\varepsilon}_i J_i)^2 + 
     \sum_{\substack{i,j\\i \ne j}} \bm{\varepsilon}_i \bm{\varepsilon}_j J_i J_j\} \\
     &= \underset{\mathbf{x}}{\mathbb{E}} \sum_{i} J_i^2 \underset{\bm{\varepsilon}}{\mathbb{E}} \{\bm{\varepsilon}_i^2\}  
     = \sigma^2 \underset{\mathbf{x}}{\mathbb{E}} \{ \|J\|^2 \}.
\label{eq:jacobian}
\end{aligned}
\end{equation}

For the second part $\mathcal{R}_H$, we have
\begin{equation}
\begin{aligned}
    \mathcal{R}_H &= \underset{\mathbf{x},\bm{\varepsilon}}{\mathbb{E}} \| \frac{1}{2} \bm{\varepsilon}^T H \bm{\varepsilon} \|^2 \\
     &= \frac{1}{4} \underset{\mathbf{x}}{\mathbb{E}} \underset{\bm{\varepsilon}}{\mathbb{E}} \{ \sum_{i,j}\bm{\varepsilon}_i^2 \bm{\varepsilon}_j^2 H_{ii} H_{jj} + \sum_{\substack{i,j\\i \ne j}} \bm{\varepsilon}_i^2 \bm{\varepsilon}_j^2 H_{ij}^2 \} \\
     &= \frac{\sigma^4}{4}\underset{\mathbf{x}}{\mathbb{E}}\{(\sum_{i}H_{ii})^2 + \|H\|_F^2-\sum_{i}H_{ii}^2\} \\
     &= \frac{\sigma^4}{4}\underset{\mathbf{x}}{\mathbb{E}}\{ \|\mathrm{Tr}(H)\|^2 + \|(\mathbf{1}-\mathbf{I}) \circ H\|_F^2 \}.
\label{eq:hessian}
\end{aligned}
\end{equation}

For the third part $\mathcal{R}_{JH}$, we have
\begin{equation}
\begin{aligned}
    \mathcal{R}_{JH} &= \underset{\mathbf{x},\bm{\varepsilon}}{\mathbb{E}} \{J \bm{\varepsilon} \bm{\varepsilon}^T H \bm{\varepsilon}\} \\
    &= \underset{\mathbf{x},\bm{\varepsilon}}{\mathbb{E}}\{(\sum_{i} \bm{\varepsilon}_i J_i) (\sum_{i,j} \bm{\varepsilon}_i \bm{\varepsilon}_j H_{ij})\}\\
    &= \underset{\mathbf{x}}{\mathbb{E}} \sum_{i,j,k} J_k H_{ij} \underset{\bm{\varepsilon}}{\mathbb{E}} \{ \bm{\varepsilon}_i \bm{\varepsilon}_j \bm{\varepsilon}_k \} 
    =0.
\label{eq:JH}
\end{aligned}
\end{equation}
Substituting Eqs.~\ref{eq:jacobian},\ref{eq:hessian} and \ref{eq:JH} into Eq.~\ref{eq:main2} completes the proof.
\end{proof}


\noindent \textbf{Analysis.} Eq.~\ref{eq:claim} characterizes our connection with the Tikhonov regularizer, where the proposed noise stability regularization has the effect of constraining the first and second-order input derivatives of the objective function $f$. 

In an analogy with previous works, our method involves common terms of the $L^2$ norm of Jacobian, the $L^2$ norm of Hessian trace, and the Frobenius norm of Hessian. Compared with \cite{bishop1995training} and \cite{rifai2011adding}, our method is capable of inheriting their merits and overcoming their flaws. Specifically, the proposed LNSR regularizes the positive guaranteed Hessian trace that avoids undesirable solutions as \cite{bishop1995training}. Compared with~\cite{rifai2011adding}, our method is more efficient, as adding noise to the input does not introduce much computation for DNNs. However, \cite{rifai2011adding} involves the gradient of Jacobian during the back-propagation process, which is considerably more complex for DNNs.

Furthermore, it is important to note that the conclusions drawn in \cite{bishop1995training} and \cite{rifai2011adding} primarily rely on the assumption of a regression task with the Mean Squared Error (MSE) loss. These approaches utilize labels to impose regularization by ensuring the perturbed input aligns with its corresponding label or minimizing the perturbed Jacobian, which also depends on labels. Although our LNSR method shares a similar concept of directly adding noise to the input, it introduces a regularizer in an unsupervised manner, leveraging the clean output as virtual supervision. As a result, our framework offers significantly greater flexibility by reducing the reliance on labels and specific forms of the loss function.

\begin{figure}[t]
    \centering
    \includegraphics[width=0.8\linewidth]{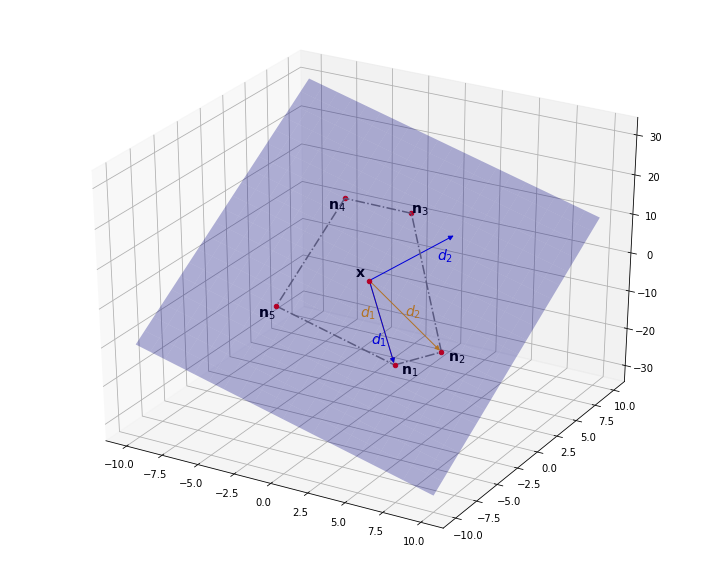}
    \caption{Illustration of a locally linear patch in a 2-dimensional manifold formed by a data point $\mathbf{x} \in \mathbb{R}^3$ and its neighbors $\mathbf{n}_1,\mathbf{n}_2,\mathbf{n}_3,\mathbf{n}_4, \mathbf{n}_5 \in \mathbb{R}^3$. Orange $\mathbf{d}_1$ and $\mathbf{d}_2$ refer to two difference vectors and blue $\mathbf{d}_1$ and $\mathbf{d}_2$ are their corresponding orthogonal vectors. Blue $\mathbf{d}_1$ and $\mathbf{d}_2$ are used to sample in-manifold noise $\bm{\varepsilon}$. The perturbed data $\mathbf{x} + \bm{\varepsilon}$ still lies in the manifold. }
    \label{fig:patch}
\end{figure}

\subsection{Analysis of In-manifold LNSR}
In this part, we shall analyze the effect of In-manifold LNSR. The main conclusion is summarized as follows. 

\noindent \textbf{Claim 2.} \emph{ Suppose that the input data lie in a k-dimensional smooth manifold, with each data point $\mathbf{x} \in \mathbb{R}^d$ and its neighbors lying on a locally linear patch. Provided there are sufficient neighbors with uniformly distributed directions around $\mathbf{x}$, noise $\bm{\varepsilon}$ sampled according to Algorithm~\ref{algo:frame} is approximate standard multivariate Gaussian in the manifold space around $\mathbf{x}$.}

Next, we provide a detailed analysis to verify the claim. First, in Section 3.4.1, we explain why the generated noise lies in the manifold, adopting the manifold assumption of Locally Linear Embedding (LLE)~\cite{roweis2000nonlinear}. Then, in Section 3.4.2, we prove that the generated noise follows the standard multivariate Gaussian distribution in the manifold space. Finally, in Section 3.4.3, we discuss the assumptions and approximations involved in the claim. 

\subsubsection{Noise on the Locally Linear Patch}
We first give a brief overview of Locally Linear Embedding (LLE)~\cite{roweis2000nonlinear} for manifold learning. Note that our purpose is not designing or apply a manifold learning algorithm. Instead, we intend to borrow the assumptions and understandings about the manifold to generate in-manifold noise. 

\noindent \textbf{Locally Linear Embedding.} The intuition behind LLE~\cite{roweis2000nonlinear} is to regard a smooth manifold as a collection of overlapping linear patches, provided these patches are small enough. Then, the local geometry can be characterized by a weight matrix $W$, which is to be solved by minimizing the reconstruction error
\begin{equation}
    R(W)=\sum_{i}\|\mathbf{x}^{(i)}-\sum_{j}W_{ij}\mathbf{x}^{(j)}\|^2.
\end{equation}
We ignore the constraints used for computing $W$ as it's not directly related to our work. Ideally, there exists an appropriate $W$ that makes $R(W)$ near zero. In such cases, each $\mathbf{x}^{(i)}$ can be approximately represented as a linear combination of its neighbors, i.e. they lie on a linear subspace. 

\noindent \textbf{In-manifold Noise.} Here we show that the noise w.r.t. a data point $\mathbf{x}$ generated according to Algorithm~\ref{algo:frame} lies on the linear patch expanded by $\mathbf{x}$ and its neighbors. It's obvious that the difference vectors $\{\mathbf{d}^{(j)}|\mathbf{d}^{(j)}=\mathbf{x}^{(j)}-\mathbf{x}\}$ lie on the same linear patch with $\mathbf{x}$ and its neighbors. Linear transformations of these $\{\mathbf{d}^{(j)}\}$, e.g. the orthogonal variants $\{\check{\mathbf{d}}^{(j)}\}$, should be still in the same subspace and, so are the linear combination of these $\{\check{\mathbf{d}}^{(j)}\}$. Therefore, the perturbed input $\mathbf{x} + \bm{\varepsilon}$ with the noise $\bm{\varepsilon}=\sum_{j=1}^{k}\varepsilon^{(j)} \check{\mathbf{d}}^{(j)}$ lies on the linear space formed by $\mathbf{x}$ and its neighbors. Note that, this does not mean a guarantee that $\mathbf{x} + \bm{\varepsilon}$ lies \emph{within} the linear patch unless both the noise magnitude and direction are properly constrained. However, by suggesting that there exist, sufficient neighbors which have diverse directions, $\mathbf{x} + \bm{\varepsilon}$ should be on the linear patch with a high probability. See Figure~\ref{fig:patch} for an intuitive illustration of the locally linear patch.

\subsubsection{In-manifold Standard Multivariate Gaussian Noise}
Recall that the noise is generated based on orthogonal difference vectors as $\bm{\varepsilon}=\sum_{j=1}^{k}\varepsilon^{(j)} \check{\mathbf{d}}^{(j)}$. When being projected to the manifold space by $M: \mathbb{R}^d \to \mathbb{R}^k$, their angles will also be preserved according to the locally linear patch assumption~\cite{roweis2000nonlinear}. So, the projected difference vectors $M(\check{\mathbf{d}}^{(j)}) \in \mathbb{R}^k$ are still orthogonal. 

Imagine that we change the coordinate system in order to let $M(\check{\mathbf{d}}^{(j)})$ be the one-hot vector where only the $j$-th element is 1 and all remaining are 0. As a result, the $j$-th element of $M(\bm{\varepsilon})=\sum_{j=1}^{k}\varepsilon^{(j)} M(\check{\mathbf{d}}^{(j)})$ is just $\varepsilon^{(j)}$, which follows the standard Gaussian distribution. Thus, the projected noise $M(\bm{\varepsilon})$ conforms the standard multivariate Gaussian distribution.


\subsection{Choices of Hyperparameters}
\label{sec:hyperparameter}
Our method involves additional hyperparameters. Here we describe how we choose a reasonable hyperparameter configuration to ensure the performance of our method. 

\subsubsection{Noise Magnitude}
We adopt an adaptive scheme to determine the magnitude of injected noise. Specifically, a noise is firstly sampled from a standard multivariate Gaussian distribution and then re-scaled by a scalar coefficient $\eta$. $\eta$ is set to be a fixed proportion between the magnitude of the original feature vector $\bm{x}$ and that of the noise vector $\epsilon$ as $\eta = 0.05 \|\bm{x}\|^2_2 / \|\bm{\varepsilon}\|^2_2$. 
\subsubsection{Noise Injected Position}
In this paper, the noise is always injected at the lowest layer, i.e. the word embedding layer, for the regularization of noise stability. In this way, we achieve the effect of stabilizing all transformer layers. Moreover, for In-manifold LNSR, obtaining hidden-layer representations of k-nearest neighbors is much more laborious since it needs additional feed-forward computations. 
\subsubsection{Number of Nearest Neighbors for In-manifold LNSR}
Under the manifold assumption, the number of nearest neighbors $b$ depends on the underlying manifold dimension. However, estimating such a dimension is often intractable for real-world data~\cite{lin2008riemannian}. In this work, we empirically find that $b=10$ usually performs well for in-manifold LNSR. While further increasing $b$ tends to violate the manifold assumption, a too-small $b$ induces an over-constrained noise space and would reduce the effectiveness of the regularization.  

\subsection{Computational Complexity}
\label{sec:complexity}

Here we provide an analysis of the computational complexity of our proposed LNSR method in comparison to relevant adversarial-based approaches. As the training process is common among these methods, our focus is specifically on the complexity associated with noise generation.

\subsubsection{Standard LNSR}
Although the regularization term of noise stability is calculated for every layer, noise generation is performed only once at the noise input layer, for which we choose the first intermediate layer. Denoting the length of input tokens as $M$, and the dimensionality of the embeddings as $d$, generating standard Gaussian noise in a training iteration requires the complexity of $O(Md)$. Note that we do not employ expensive high-dimensional multivariate Gaussian sampling. 

\subsubsection{In-manifold LNSR}
This advanced method involves additional calculations on searching for k-nearest neighborhoods and forming a manifold space for each input token. Through the utilization of widely adopted space partitioning algorithms, the computational complexity of the first component can be expressed as $O(Mkd*\log (N))$, where $N$ is the size of the vocabulary used in K-NN searching. As for the latter component, only linear operations are needed with the complexity of $O(Mkd)$.

\subsubsection{Adversarial perturbations} 
Since adversarial-based methods aim to calculate the worst-case perturbation for a given input instance, it typically requires several training iterations over the entire network to guarantee an optimal solution regarding the adversarial objective. Denoting the number of layers as $L$, and the number of iterations by $T$, the complexity of generating adversarial perturbations will be dominated by $O(TLM^2d)$. Note that this term only accounts for the forward computation while disregarding the gradient calculation on the input.

As all the complexity approximations involve the same $d$, we ignore it for easier comparison. Given the approximate values as $M=10, k=10, N=10^5, T=10, L=10$, the complexity approximations for Standard LNSR, In-manifold LNSR, and Adversarial perturbations are $O(10d)$, $O(10^3d)$ and $O(10^3d)$, respectively. It can be observed that our Standard LNSR is much more efficient than the other two. Moreover, our LNSR approaches exhibit better scalability regarding the input sequence length.

\section{Experiments}

\subsection{Datasets}
To verify the effectiveness of our method for improving the generalizability of language models, we conduct experiments on text classification and question-answering (QA) tasks, respectively.
\subsubsection{Text Classification Tasks}
For the text classification task, we conduct experiments on four few-sample (less than 10k training samples) text classification tasks of GLUE\footnote{https://gluebenchmark.com/}, we present a brief description below and refer readers to Appendix \ref{sec:experimental} Table \ref{tab:datasets} for more details.

\textbf{Corpus of Linguistic Acceptability} (CoLA \cite{Warstadt2019NeuralNA}) is an English acceptability judgments dataset consisting of 10657 sentences from 23 linguistics publications. 
Each sentence is annotated with a binary label, indicating whether this sentence is grammatical in English. 
The task is classification and we use Matthews correlation coefficient (MCC) \cite{matthews1975comparison} as the evaluation metric.

\textbf{Microsoft Research Paraphrase Corpus} (MRPC \cite{Dolan2005AutomaticallyCA}) is a corpus for the paraphrase detection task. Each example is a sentence pair 
, whose label is 1 if the two sentences in this pair are equivalent in semantics. 
We evaluate the performance with the commonly adopted Accuracy and average F1 score. 

\textbf{Recognizing Textual Entailment} (RTE \cite{Wang2018GLUEAM}) \cite{Dagan2005ThePR} \cite{BarHaim2006TheSP} \cite{Giampiccolo2007TheTP} is a corpus for the textual entailment task. Each example is a sentence pair whose label is whether the first entails the second. The evaluation metric is Accuracy.

\textbf{Semantic Textual Similarity Benchmark} (STS-B \cite{Cer2017SemEval2017T1}) is a task for determining the semantic similarity of a sentence pair. 
The similarity is represented by integral numbers $\{1,2,3,4,5\}$, which the model is learned to predict. Common metrics for evaluation are the Pearson and Spearman correlation coefficients, and we report the average of them.

\subsubsection{Question Answering Tasks}
For the question-answering task, we use the \textbf{S}tanford \textbf{Q}uestion \textbf{A}nswering \textbf{D}ataset (SQuAD) \cite{rajpurkar2016squad} as an in-domain question answering task to evaluate our method on more complex NLP problems. Besides, we use the \textbf{M}achine \textbf{R}eading for \textbf{Q}uestion \textbf{A}nswering (MRQA) 2019 \cite{fisch2019mrqa} as an out-of-domain question answering task to verify the effectiveness of our method for improving the domain generalizability of language models. 

SQuAD is a Machine Reading Comprehension (MRC) dataset. The question-answer (QA) pairs in this dataset are constructed from Wikipedia articles by crowd workers. For each QA pair, the answer is a segment of text, or span that can be extracted from the corresponding reference passage, or the question might be unanswerable. We use the SQuAD v 1.1 dataset for experiments.

MRQA is a task for testing extractive question-answering models on their ability to generalize to the out-of-domain data. In MRQA 2019, researchers collect different extractive QA datasets and unify the format of data. The goal is to test how models trained on “in-domain” datasets can generalize to the “out-of-domain” datasets. More details about the different QA datasets are summarized in Appendix \ref{sec:experimental}.

\subsection{Baseline Models}

\textbf{Fine-tuning}. Fine-tuning refers to the vanilla language model fine-tuning method. We adopt the standard fine-tuning strategy of BERT and RoBERTa described in~\cite{Devlin2019BERTPO} and \cite{Liu2019RoBERTaAR}.

\textbf{$\text{L}^2$-SP} \cite{Li2018ExplicitIB} is a regularization scheme that is used for constraining the extent of parameters to update while fine-tuning a pre-trained model. The goal of introducing this regularization item is to 
preserve the general knowledge contained in the pre-trained model.
The form of the regularizer is  $\Omega(w)=\frac{\alpha}{2}||w_s-w_s^0||+\frac{\beta}{2}||w_{\bar{s}}||$, where $w_s$ are parameters shared by the pre-trained and fine-tuned models and $w_{\bar{s}}$ are those specific to the target task.

\textbf{Mixout} \cite{Lee2020MixoutER} is a regularization method motivated by Dropout \cite{Srivastava2014DropoutAS} and DropConnect \cite{Wan2013RegularizationON}. At each training iteration, instead of replacing parameters with 0, Mixout replaces parameters with their pre-trained value with a probability $p$.

\textbf{SMART} \cite{Jiang2020SMARTRA} is a noise-based regularization method that adopts an adversarial training strategy to promote the models' smoothness. Besides, SMART employs a Bregman proximal point optimization method to prevent the model from aggressively updating during fine-tuning.

\textbf{FreeLB} \cite{Zhu2020FreeLBEA} is a simple adversarial noise regularization method that is applied on input embeddings.

\textbf{CWGNC} \cite{yang2022improving} (\textbf{C}omponent-Wise \textbf{G}radient \textbf{N}orm \textbf{C}lipping) clips the gradient norm of the Key-Query-Value parameters individually in Transformer to balance the distribution of gradients across different components to adjust their convergence speed to improve the fine-tuning of language models.

\subsection{Experimental Setup and Implementation Details}
Our model is implemented using Pytorch based on Transformers framework\footnote{https://huggingface.co/transformers/index.html} \cite{Wolf2019HuggingFacesTS} and the backbone language models are BERT \cite{Devlin2019BERTPO} and RoBERTa \cite{Liu2019RoBERTaAR}. We adopt settings of learning strategies and hyperparameters recommended by Devlin et al. \cite{Devlin2019BERTPO}. We use the Huggingface edition AdamW \cite{Kingma2015AdamAM} optimizer with a learning rate $\in \{2 \times 10^{-5}, 3 \times 10^{-5}, 5 \times 10^{-5} \} $ and a batch size $\in \{16, 32, 64\}$, and the $\beta_1=0.9,\ \beta_2=0.999$. We adopt the regularization weight in the range of $\{1.0, 0.8, 0.6, 0.4, 0.2\}$ for different tasks. The warm-up ratio for the classification task and question-answering task are set to 6\% and 10\% respectively. For a fair comparison, we set the maximum numbers of epochs to 3 and 2 for classification and question-answering tasks respectively, which is the same as baseline modes' experiment settings described in the related papers. For the hyperparameters used in the two LNSR methods, we use those described in Section~\ref{sec:hyperparameter}. We employ Faiss \cite{JDH17} for an efficient implementation of k-NN search to generate in-manifold noise.
\vspace{-0.8cm}

\begin{figure*}[htbp]
  \centering
  \includegraphics[width=0.9\linewidth]{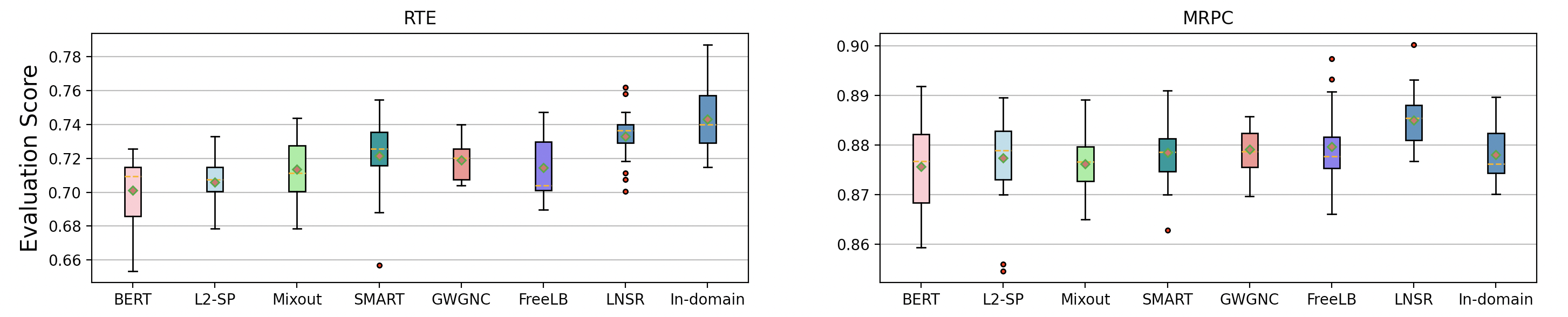}
  \includegraphics[width=0.9\linewidth]{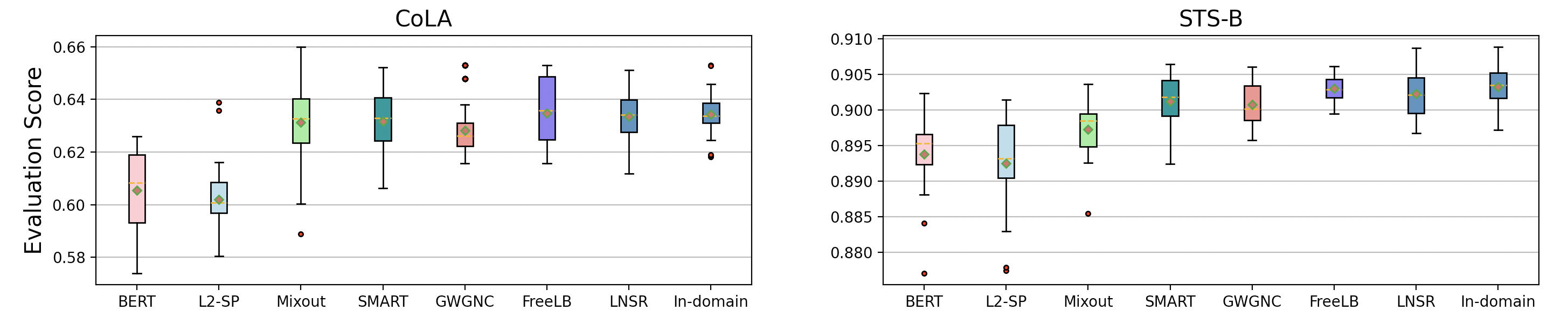}
 \caption{Performance distribution of different models on the selected GLUE benchmark across 25 random seeds.}
\label{fig:overall}
\end{figure*}
\begin{center}
\begin{table*}[htbp]
  \resizebox{\linewidth}{!}{
  \begin{tabular}{lllllllllllll}
    \toprule
     \multirow{2}{*}{}& \multicolumn{3}{c}{RTE} & \multicolumn{3}{c}{MRPC} & \multicolumn{3}{c}{CoLA}& \multicolumn{3}{c}{STS-B} \\
\cmidrule(r){2-4} \cmidrule(r){5-7} \cmidrule(r){8-10} \cmidrule(r){11-13}
\makecell[l]{$\text{BERT}_{\text{LARGE}}$} &  mean      &  std   $\downarrow$&   max
&  mean      &  std  $\downarrow$ &   max
&  mean      &  std  $\downarrow$ &   max 
&  mean      &  std  $\downarrow$ &   max \\

\midrule
   FT \cite{Devlin2019BERTPO}& $70.13$& $1.84$& $72.56 $& $87.57/89.75$&$0.92$&$89.16/91.12 $&$60.54 $&$1.49 $& $62.59$ &$89.38 $&$0.53 $& $90.23$ \\
    $\text{L}^2\text{-SP}$ \cite{Li2018ExplicitIB}& $70.58$& $1.29$& $73.28$& $ 87.74/89.84$&$0.86 $&$88.95/90.90 $& $60.19$ &$1.42 $&$63.89 $& $89.25$ & $0.62$&$90.14 $\\
    Mixout\cite{Lee2020MixoutER}& $71.35$& $1.66$& $74.36$& $87.63/89.80$&$0.62$&$88.91/90.81 $&$63.12 $&$1.68 $& $65.12$ &$89.58 $&$0.35 $& $90.11$ \\
    SMART\cite{Jiang2020SMARTRA}& $72.23$& $2.41$& $75.45$& $87.86/89.80$&$0.63$&$89.09/90.94 $&$63.16 $&$1.17 $& $65.21$ &$90.11 $&$0.33 $& $90.83$ \\
    CWGNC\cite{yang2022improving}& $71.94$& $\bf{1.02}$& $74.01$& $87.93/90.09$&$0.62$&$88.49/91.85 $&$63.00 $&$1.06 $& $65.30$ &$90.10 $&$0.32 $& $90.60$ \\
    FreeLB\cite{Zhu2020FreeLBEA}& $71.87$& $1.87$& $74.47$& $87.88/89.97$&$0.78$&$88.92/90.79 $&$63.42 $&$1.37 $& $65.79$ &$90.28 $&$\bf{0.28} $& $90.67$ \\
    \midrule
    \bf{Standard LNSR} & ${73.31}$& ${1.55}$& $76.17$& $\bf{88.50/90.42}$&$\bf{0.56}$&${\bf90.02}/91.81 $&$63.35 $&$1.05 $& $\bf{65.99}$ &$90.23 $&$0.31 $& $90.97$\\
    \textbf{In-manifold LNSR} & $\bf{74.33}$& $1.78$& $\bf{78.70}$& $87.92/89.21$&$0.62$&$88.97/\bf{92.31}$&$\bf{63.44} $&$\bf{1.01} $& $65.30$ &$\bf{90.32} $&$0.29 $& $90.88$\\
    \bottomrule
  \end{tabular}
 }
  \resizebox{\linewidth}{!}{
  \begin{tabular}{lllllllllllll}
    \multirow{2}{*}{} \\
\makecell[l]{$\text{RoBERTa}_\text{{LARGE}}$} \\
    \midrule
    FT \cite{Liu2019RoBERTaAR}& $80.99$& $2.96$& $85.19 $& $88.51/91.72$&$0.90$&$90.69/93.17 $&$65.27 $&$1.95 $& $67.74$ &$91.29 $&$0.40 $& $92.21$ \\
    $\text{L}^2\text{-SP}$ \cite{Li2018ExplicitIB}& $82.47$& $1.59$& $85.56$& $ 88.52/91.67$&$\bf{0.54}$&$89.46/92.52 $& $64.40$ &$1.03 $&$66.54 $& $91.83$ & $0.25$&$92.38$\\

    Mixout\cite{Lee2020MixoutER}& $81.76$& $2.67$& $84.84$& $88.57/91.79$&$0.73$&$89.95/92.94 $&$64.68 $&$1.53 $& $67.26$ &$91.93 $&$0.25 $& $92.35$ \\
    SMART\cite{Jiang2020SMARTRA}& $82.40$& $1.78$& $85.55$& $88.64/91.85$&$0.98$&$90.19/92.93 $&$65.81 $&$1.19$& $67.74$ &$91.89 $&$0.19 $& $92.18$ \\
    CWGNC\cite{yang2022improving}& $82.49$& $\bf{1.36}$& $85.19$& $88.88/91.92$&$0.76$&$89.95/92.66$&$65.86 $&$\bf{0.84} $& $67.97$ &$91.85 $&$0.23 $& $92.16$ \\
    FreeLB\cite{Zhu2020FreeLBEA}& $82.85$& $2.35$& $85.92$& $89.67/91.30$&$0.59$&$90.68/93.33 $&$64.88 $&$1.17 $& $67.76$ &$91.85 $&$0.18 $& $92.05$ \\
    \midrule
    \bf{Standard LNSR} & $81.48$& $1.49$& $83.76$& $88.65/91.72$&$0.86$&$90.93/93.16 $&$65.32 $&$1.25 $& $67.49$ &$91.96 $&$\bf{0.17} $& $92.24$\\
    \textbf{In-manifold LNSR}& $\bf{82.92}$& $1.91$& $\bf{86.64}$& $\bf{90.25/92.13}$&$0.85$&$\bf{91.91/93.55} $&$\bf{65.92} $&$1.47 $& $\bf{68.71}$ &$\bf{92.03}$ &$0.28 $ & $\bf{92.59}$\\
    \bottomrule
  \end{tabular}
  }
    \caption{The mean/max evaluation scores and standard deviation values on the selected GLUE benchmark datasets across 25 random seeds when fine-tuning the BERT and the RoBERTa models with different regularization methods. The evaluation metrics of MRPC are Acc/F1.}
    \label{tab:overall}
\end{table*}
\end{center}

\subsection{Overall Performance}
For the text classification task, Table~\ref{tab:overall} shows the performance of different models on selected GLUE datasets. Each dataset is trained over 25 random seeds. In the LNSR method, we uniformly inject noise at the first layer on BERT-large and RoBERTa-large for comparison with baseline models. For the In-manifold LNSR, we adopt different mix ratios for different tasks \{RTE:0.1, MRPC:0.12, CoLA:0.15, STSB:0.2 \}. As we can see from the table, pre-trained language models with LNSR and In-manifold LNSR outperform all the baseline models in mean and max values, which indicates the stronger generalizability of our model over other baseline models. To verify whether the performance gains of our methods are significant, we calculate the p-values between the performance distributions of the fine-tuning baseline and our proposed LNSR methods. 
We get very small p-values on all tasks that, $9.7\times10^{-7}$ on RTE, $2.3\times10^{-4}$ on MRPC, $4.7\times10^{-8}$ on CoLA, and $3.3\times10^{-8}$ on STS-2. 

Standard deviation can be used to reflect the stability of a learning procedure. 
In this work, a higher std means that the model is more sensitive to random seeds. According to the results of experiments, 
models with LNSR have relatively low standard deviations on most tasks, suggesting that our method is less sensitive to randomness involved by data orders and initializations. 
In addition, Figure~\ref{fig:overall} provides a clearer illustration. Although the In-manifold LNSR methods have a higher std deviation compared with LNSR, it gains more improvement on mean and max values. In summary, the two proposed  LNSR methods 
can not only improve the average performance but also reduce the instability of BERT fine-tuning.  

\begin{figure}[htbp]
  \centering
  \includegraphics[width=0.99\linewidth]{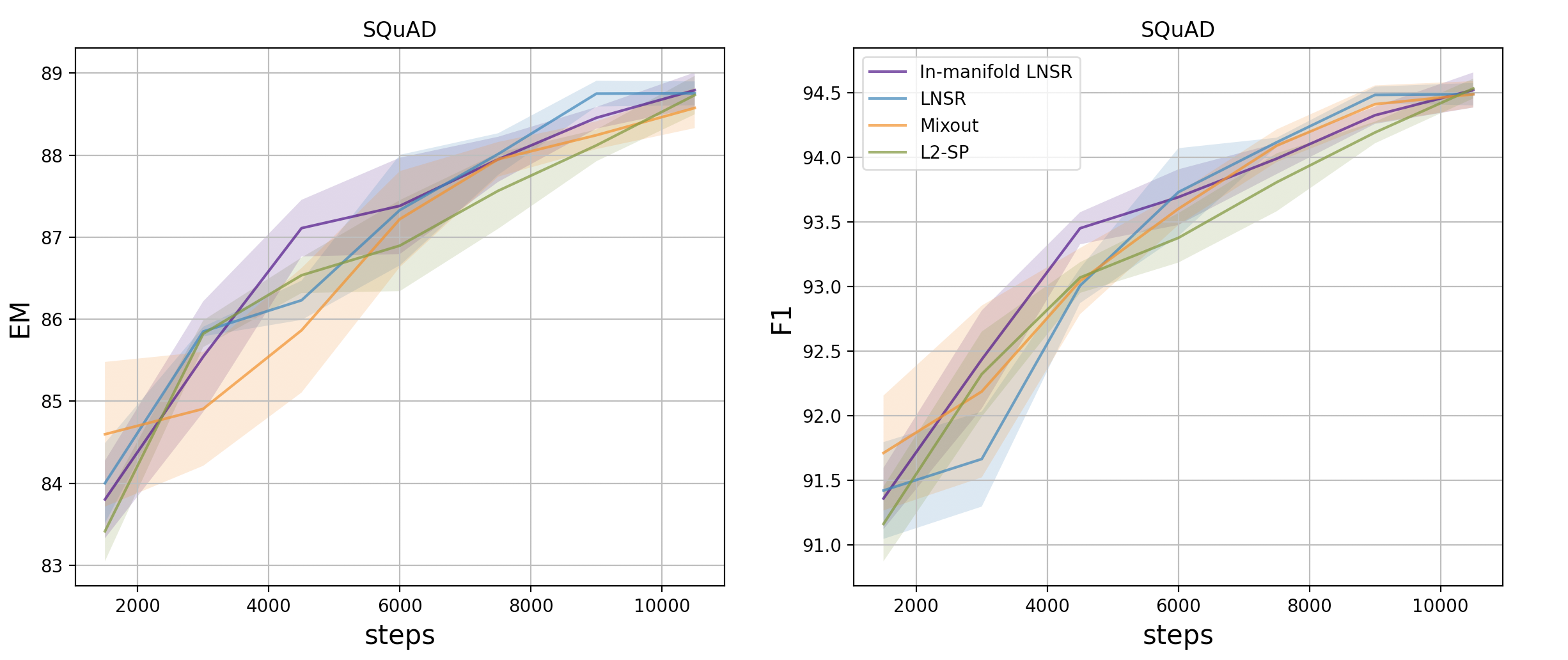}
 \caption{The mean (solid lines) and range (shaded
region) of EM and F1 scores during fine-tuning RoBERTa-Large on SQuAD datasets, across 5 random seeds.}
\label{fig:conv}
\end{figure}

For the Question Answering task, Table~\ref{tab:qaoverall} shows the results of all the methods on the SQuAD dataset. We can see that on the more challenging question-answering task with a larger dataset, our proposed LNSR and In-manifold LNSR methods can still improve the models' fine-tuning performance compared with other methods. Specifically, the BERT large model fine-tuning with In-manifold LNSR achieves an average and max dev-set EM/F1 of 86.88/93.07 and 87.48/93.40; the RoBERTa large model fine-tuning with In-manifold LNSR achieves an average and max dev-set EM/F1 of 88.71/94.43 and 88.93/93.40. Figure~\ref{fig:conv} shows the mean and the range of EM/F1 scores' changes during the fine-tuning process. \footnote{Due to the high training cost, we adopt 5 random seeds for comparison.} The p-values between the performance
distributions of the fine-tuning baseline and our proposed LNSR are $3.2\times10^{-8}$/$1.9\times10^{-8}$, and for the In-manifold LNSR, the p-values are $4.1\times10^{-8}$/$2.7\times10^{-8}$.
\begin{center}
\begin{table}[h!]
  \label{tab:commands}
  \resizebox{\columnwidth}{!}{
  \begin{tabular}{lllllll}
    \toprule
    \multirow{2}{*}{}& \multicolumn{3}{c}{EM} & \multicolumn{3}{c}{F1} \\
\cmidrule(r){2-4} \cmidrule(r){5-7} \makecell[l]{$\text{BERT}_{\text{LARGE}}$} &  mean      &  std   $\downarrow$&   max
&  mean  &  std   $\downarrow$&   max \\
\midrule
    FT \cite{Devlin2019BERTPO}& $86.75$& $0.15 $& $86.95 $& $92.95$& $0.08 $& $93.06 $\\
    $\text{L}^2\text{-SP}$ \cite{Li2018ExplicitIB}&$86.87$ &$\bf{0.09}$ &$86.98$ &$93.06$&$\bf{0.06}$&$93.15$ \\
    Mixout\cite{Lee2020MixoutER}&$86.81$&$0.24$&$87.08$&$93.04$&$0.13$ &$93.18$ \\
    SMART\cite{Jiang2020SMARTRA}&$86.81$&$0.24$&$87.19$&$92.99$
    &$0.10$ &$93.17$ \\
    \midrule
    \bf{Standard LNSR} &$86.88$&$0.11$&$87.15$&$93.06$&$0.10$&$93.18$\\
    \textbf{In-manifold LNSR}& $\bf{86.95}$&$ 0.22$ &$\bf{87.48}$& $\bf{93.07}$&$ 0.14$ &$\bf{93.40}$ \\
    \bottomrule
  \end{tabular}
  }
  \resizebox{\columnwidth}{!}{
  \begin{tabular}{lllllll}
    \multirow{2}{*}{}\\
\makecell[l]{$\text{RoBERTa}_{\text{LARGE}}$} \\
\midrule
    FT \cite{Devlin2019BERTPO}& $88.47$& $0.21$& $88.82$& $94.26$& $0.11$ & $94.46 $\\
    $\text{L}^2\text{-SP}$ \cite{Li2018ExplicitIB}&$88.86$&$\bf{0.12}$&$88.97$&$94.53$&$0.08$&$94.57$ \\
    Mixout\cite{Lee2020MixoutER}&$88.59$&$0.18$&$88.85$&$94.42$&$0.11$&$94.57$\\
    SMART\cite{Jiang2020SMARTRA}&$88.69$&$0.17$&$88.91$&$94.38$
    &$\bf{0.05}$ &$94.48$\\
    \midrule
    \bf{Standard LNSR} &$88.71$&$0.18$&$88.88$&$94.43$&$0.13$ &$94.63$\\
    \textbf{In-manifold LNSR}& $\bf{88.93}$&$0.13$ &$\bf{89.09}$& $\bf{94.56}$&$ 0.12$ &$\bf{94.72}$\\
    \bottomrule
  \end{tabular}
  }
    \caption{The mean/max evaluation scores and standard deviation values on the SQuAD dataset across 5 random seeds when fine-tuning the BERT and the RoBERTa models with various regularization methods.}
    \label{tab:qaoverall}
\end{table}
\end{center}

\section{Analysis}
\subsection{Resilience to Domain Shift}
To verify the robustness of the domain shift of our methods, we investigate the domain generalization performance of different methods on the MRQA 2019 benchmark. We first fine-tune a language model on SQuAD and then evaluate the fine-tuned model on the out-of-domain datasets.
 
As shown in Table~\ref{tab:DG}, language models fine-tuned with our LNSR methods outperform the vanilla fine-tuning method overall on the mean and max F1 scores on each of the MRQA out-of-domain datasets. In addition, compared with SMART, our methods obtain a better F1 score on most datasets which shows the better generalizability of our method. In-manifold LNSR performs better than the vanilla LNSR on most datasets. The overall results demonstrate our LNSR methods are more robust to domain shift problems, and models fine-tuned with our methods show a better ability for zero-shot domain transfer. For this observation, we argue that LNSR can promote the generalization capacity of fine-tuned language models so that the models show higher resilience to domain shift. 

\begin{center}
\begin{table}[htbp]
  \label{tab:commands}
  \resizebox{\columnwidth}{!}{
  \begin{tabular}{ll|llll}
    \toprule
\makecell[l]{Dataset}& \makecell[l]{Domain} &\makecell[l]{FT}&\makecell[l]{SMART}&\makecell[l]{\bf Standard LNSR}&\makecell[l]{\bf In-manifold LNSR}
\\
\midrule
SQuAD &Wiki&$94.26/94.46 $&94.39/94.47&$94.43/94.63$&$
{\bf{94.56}}/{\bf{94.72}}$\\
   \midrule

   
   DROP &Wiki&$57.63/59.37 $&58.49/60.11&${\bf{58.79}}/ {\bf{61.1}}$&$58.13/59.83$\\
   DuoRC &Movie&$61.83/62.98 $&62.31/63.55&$61.99/63.01$&$\bf62.42/63.72$\\
   RE & Wiki&$88.02/88.27 $&88.10/88.49&$88.23/88.58$&$\bf88.26/88.86$\\
   RACE &Exam&$52.48/53.78 $&{\bf{53.34}}/53.81&${52.95}/53.51$&$52.82/\bf53.87$\\
   TextbookQA &Book&$51.86/55.33 $&{\bf53.98/56.54}&$52.90/55.10$&${53.64}/55.76$\\
   BioASQ & Bio&$65.58/66.27 $&65.26/66.04&$66.40/67.02$&${\bf67.01}/\bf67.88$\\
    \bottomrule
  \end{tabular}
 }
    \caption{F1 mean and max values for models trained on
SQuAD and evaluated on out-of-domain datasets from
the MRQA 2019 shared task, across 5 random seeds.}
    \label{tab:DG}
\end{table}
\end{center}
\begin{center}
\begin{table*}
  \label{tab:commands}
  \resizebox{\textwidth}{!}{
  \begin{tabular}{llllllllllllll}
    \toprule
\multirow{2}{*}{}& \multicolumn{3}{c}{RTE} & \multicolumn{3}{c}{MRPC} & \multicolumn{3}{c}{CoLA}& \multicolumn{3}{c}{STS-B} \\
\cmidrule(r){2-4} \cmidrule(r){5-7} \cmidrule(r){8-10} \cmidrule(r){11-13}
\makecell[l]{$\text{RoBERTa}_{\text{LARGE}}$} &  mean      &  std   $\downarrow$&   max
&  mean      &  std  $\downarrow$ &   max
&  mean      &  std  $\downarrow$ &   max 
&  mean      &  std  $\downarrow$ &   max \\
\midrule
    FT \cite{Liu2019RoBERTaAR}& $80.99$& $2.96$& $85.19 $& $88.51/91.72$&$0.90$&$90.69/93.17 $&$65.27 $&$1.95 $& $67.74$ &$91.99 $&$0.40 $& $92.50$ \\
    \midrule
    Mix-ratio: 0.10& 
    $82.10$&$0.79$&$84.11$& $88.89/91.98$&$0.67$&$90.44/93.12 $&
    $65.03 $&$1.24 $&$68.52$ &
    $91.93 $&$0.30 $& $92.32$\\
    Mix-ratio: 0.12&
    $81.26$&$1.30$&$84.12$& $88.64/91.80$&$0.70$&$90.20/92.81 $&
    $65.13 $&$1.24 $&$67.25$ &
    $92.07 $&$0.18 $& $92.50$\\
    Mix-ratio: 0.15&
    $80.88$&$1.60$&$84.48$& 
    $88.58/91.76$&$0.63$&$89.71/92.81 $&
    $65.63 $&$1.16 $&$67.98$ &
    $92.00 $&$0.18 $& $92.39$\\
    Mix-ratio: 0.20&
    $81.51$&$1.58$&$84.12$& 
    $88.17/91.49$&$1.80$&$89.71/92.53 $&
    $65.99 $&$1.46 $&$70.20$ &
    $92.09 $&$0.19 $& $92.55$\\
    \bottomrule
  \end{tabular}
  }
    \caption{The mean/max evaluation scores and standard deviation values on the selected GLUE benchmark datasets across 25 random seeds when fine-tuning the RoBERTa model with different In-manifold noise scales.}
    \label{tab:mgeng}
\end{table*}
\end{center}

\begin{center}
\begin{table*}
  \label{tab:commands}
  \resizebox{\textwidth}{!}{
  \begin{tabular}{lllllllllllll}
    \toprule
    \multirow{2}{*}{} & \multicolumn{3}{c}{RTE} & \multicolumn{3}{c}{MRPC} & \multicolumn{3}{c}{CoLA}& \multicolumn{3}{c}{STS-B} \\
\cmidrule(r){2-4} \cmidrule(r){5-7} \cmidrule(r){8-10} \cmidrule(r){11-13} \makecell[l]{$\text{BERT}_{\text{LARGE}}$}
&  mean      &  std $\downarrow$  &   max
&  mean      &  std $\downarrow$  &   max
&  mean      &  std $\downarrow$  &   max 
&  mean      &  std $\downarrow$  &   max \\
\midrule
    FT & $70.13$& $1.84$& $72.56 $& $87.57/89.75$&$0.92$&$89.16/91.12 $&$60.54 $&$1.49 $& $62.59$ &$89.38 $&$0.53 $& $90.23$ \\
    FT (4 Epochs) & $70.69$& $1.97$& $73.65 $& $88.15/90.20$&$0.65$&$89.21/91.15 $&$60.69 $&$1.24 $& $62.09$ &$89.29 $&$0.56 $& $90.12$ \\
    FT+Gaussian Noise & $70.62$& $1.56$& $72.93 $& $87.95/90/15$&$0.83$&$89.33/91.16 $&$60.18 $&$1.58 $& $62.59$ &$89.34 $&$0.51 $& $90.11$ \\
    FT+In-manifold Noise & $73.14$& $2.11$& $77.97 $& $87.41/90.31$&$0.67$&$88.24/91.72 $&$61.62 $&$1.76 $& $64.32$ &$90.07 $&$0.81 $& $90.75$\\
    \bf{Standard LNSR} & ${73.31}$& $\bf{1.55}$& $76.17$& $\bf{88.50/90.42}$&$\bf{0.56}$&${\bf90.02}/91.81 $&$63.35 $&$1.05 $& $\bf{65.99}$ &$90.23 $&$0.31 $& $\bf90.97$\\
    \textbf{In-manifold LNSR}& $\bf{74.33}$& $1.78$& $\bf{78.70}$& $87.92/89.21$&$0.62$&$88.97/\bf{92.31}$&$\bf{63.44} $&$\bf{1.01} $& $65.30$ &$\bf{90.32} $&$\bf{0.29} $& $90.88$\\
    \bottomrule
  \end{tabular}
  }
    \caption{Ablation study of different fine-tuning methods on the selected GLUE benchmark datasets, we report the mean/max evaluation scores and standard deviation values across 25 random seeds.}
    \label{tab:ablation}
\end{table*}
\end{center}

\subsection{Ablation Study}
We conduct ablation experiments on text classification tasks to further validate the mechanism of both the standard and In-manifold LNSR methods. We compare our method with fine-tuning more epochs of the BERT model and injecting noise without an explicit regularization (we add Gaussian/In-manifold noise to the intermediate representation of a BERT layer in the forward propagation process without explicitly regularizing the noise stability), respectively. The results are in Table ~\ref{tab:ablation}. We observe that the performance gains brought by more training epochs are less significant. Meanwhile, fine-tuning by only imposing perturbations can not bring satisfying improvement in performance too. However, fine-tuning with our proposed LNSR methods achieves satisfying results on every task, which adds to the mounting evidence that fine-tuning with LNSR can effectively improve the generalizability of language models.

\vspace{-0.2cm}
\subsection{Sensitivity to the Position of Noise Injection in LNSR}
\label{sec:sensitivity}
As illustrated in Figure~\ref{fig:noise}, the performance of BERT fine-tuning is sensitive to the layer of noise injection. So we investigate the impact of different positions of noise injection on BERT fine-tuning. For comparison, we inject noise into different layers of the BERT model and regularize the noise stability item. According to the experimental results shown in Figure \ref{fig:sens} in Appendix \ref{sec:appc}, we can conclude that all injection positions bring significant improvements over vanilla fine-tuning. Particularly, noise injected into the lower layers usually brings more performance gains of the BERT fine-tuning, which indicates that lower-layer LNSR may be more effective as it influences more layers (i.e. parameters). 
\vspace{-0.2cm}
\subsection{Sensitivity to the Mix-ratio of In-manifold LNSR}
To investigate the influence of the mix-ratio of In-manifold LNSR on language model fine-tuning, we conduct experiments on RoBERTa with different scales of In-manifold noise on different text classification tasks. To avoid catastrophic changes in the embedding layer, the noise scale should be reasonably restricted. Specifically, we evaluate choices of \{0.10, 0.12, 0.15, 0.20\}. 
As we can see from Table~\ref{tab:mgeng}, the performance of In-manifold LNSR fine-tuning is affected by both the scale of datasets and the scale of noise mix-ratio. For the RTE dataset with 2.5K training samples, we obtain the best mean/max value of 82.10/84.11 under the mix-ratio of 0.1, while larger mix-ratios can lead to an unstable and even collapsed fine-tuning process. 
But on larger datasets like MRPC, CoLA, and STS-B (which have 3.7K, 8.5K, and 7k training examples, respectively), 
larger mix ratios can bring more absolute performance gains. 

It can be concluded that, under a reasonable restriction, the best choice of the noise magnitude (at least partially) depends on the scale of training data. For a larger training set, our method tends to achieve higher performance with larger noise magnitudes. However, when fine-tuning on smaller datasets, we suggest smaller noise magnitudes to prevent unexpected representation collapse.

\subsection{Relationship to Existing Relevant Work}
Our algorithm is relevant to the noise-based methods including SMART \cite{Jiang2020SMARTRA}, FreeLB \cite{Zhu2020FreeLBEA} and R3F \cite{Aghajanyan2020BetterFB}, most of which focus on the robustness of few-sample fine-tuning through a fashion of adversarial training. 
Specifically, SMART uses the gradient ascent method to learn a noise constrained within an $\epsilon$-ball and then minimize the distributional difference between the original and the perturbed representations. FreeLB proposes to directly minimize the adversarial loss  $\mathcal{L}_{FreeLB}(\theta)=\text{sup}_{\Delta\theta:|\Delta\theta|\leq\epsilon} \mathcal{L}(\theta+\Delta\theta)$, implemented by iterative gradient updates. R3F improves the efficiency by removing the procedure of adversarial optimizing in SMART and proposes to directly improve the smoothness.

Compared with this type of adversarial training-based algorithm, our LNSR not only simplifies the process of noise perturbation and reduces the computing complexity as analyzed in Section III (F), but also enjoys additional properties which are essential to model generalization.

First, while adversarial-example-based methods are motivated by worst-case robustness w.r.t. small perturbations on input data, our approach is more directly associated with statistical learning principles, i.e., the generalization bound. Specifically, \cite{arora2018stronger} figured out noise stability as a computation-tractable metric to bound the generalization error. Moreover, our approach has implicit equivalence to the Tikhonov regularizer, which is widely applied to shallow models for controlling model complexity. 

Second, our method is expected to better simulate real-world data noise compared to adversarial noise. Previous research~\cite{tsiprasrobustness} has confirmed these two kinds of robustness, i.e., to natural noise and adversarial noise, are fundamentally conflicting. Given that our work aims at natural scenarios with random data noise rather than artificial adversarial noise, the proposed LNSR is a more reasonable solution. 

\subsection{Comparison of Gaussian Noise and In-manifold Noise}
\begin{figure}[htbp]
  \centering
  \includegraphics[width=0.49\linewidth]{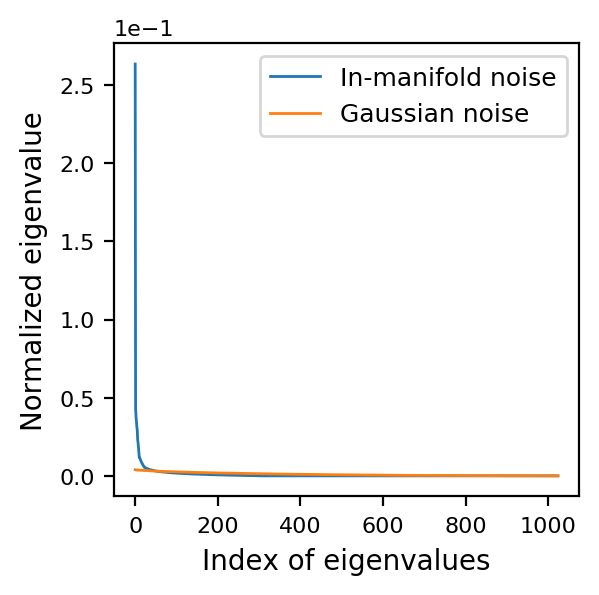}
  \includegraphics[width=0.49\linewidth]{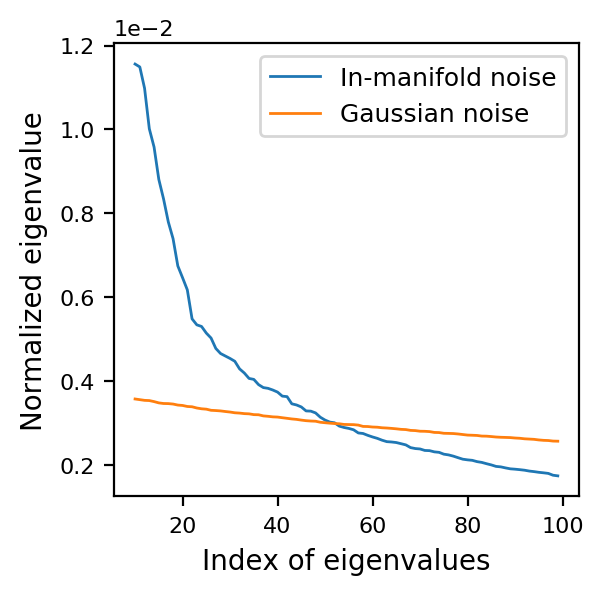}
  \caption{PCA eigenvalues of Gaussian Noise and In-manifold Noise. Eigenvalues are sorted in a descending order. For comparison, we normalize the area under the eigenvalue curve by dividing all eigenvalues by their sum. The top plot shows all eigenvalues. The bottom plot zooms in the index interval between 10 and 100.}
  \label{fig:pca}
\end{figure}

To further validate the impact of In-manifold noise, we conduct spectral decomposition analysis. Specifically, we perform Principal Components Analysis (PCA) on a batch of noise vectors sampled by the standard Gaussian distribution and the In-manifold strategy, respectively. As shown in Figure~\ref{fig:manifold}, informally, the sampled In-manifold noise has very limited freedom of direction, if data points actually lie on a low-dimensional manifold. Therefore, such a batch of noise vectors could be characterized by only a few major directions. The PCA eigenvalue distribution in Figure~\ref{fig:pca} shows a remarkable  difference between the standard and In-manifold Gaussian noise. For the In-manifold noise, almost all information is compressed on the top of a few eigenvectors, indicating its low actual dimensionality. In contrast, the standard Gaussian noise has a relatively smooth distribution of eigenvalues. Our result is consistent with recent studies about the embedding space of BERT, e.g., \cite{cai2020isotropy} points out that the word embedding in BERT usually has a local intrinsic dimension of less than 10. 

\section{Related Work}
\subsection{Pre-training}
Pre-training technology \cite{Devlin2019BERTPO,Erhan2009TheDO,Erhan2010WhyDU} has orchestrated tremendous progress in the natural language processing area in the past few years. In early NLP works, pre-training mainly focuses on using distributional representations (i.e., word embeddings) for individual words \cite{Pennington2014GloveGV,Mikolov2013DistributedRO}. Furthermore, Dai et al. \cite{Dai2015SemisupervisedSL} propose to first train a general language model 
using a self-supervised learning method and then adapt the obtained language model to downstream tasks. In recent years, large-scale pre-trained language models, such as ELMo \cite{Peters2018DeepCW}, GPT/GPT-2/GPT-3 \cite{Radford2018ImprovingLU,Radford2019LanguageMA,brown2020language}, BERT \cite{Devlin2019BERTPO}, XLNet \cite{Yang2019XLNetGA}, RoBERTa \cite{Liu2019RoBERTaAR}, ELECTRA \cite{clark2020electra}, T5\cite{2020t5}, etc. have achieved tremendous success in NLP due to the powerful ability to contextual representation and transferability to downstream tasks. In a typical pre-training and fine-tuning paradigm, language models are
first pre-trained on a large amount of unlabeled data (e.g., common crawl, C4) to capture rich semantic information of natural languages, and then adapted to the downstream tasks using labeled datasets \cite{Devlin2019BERTPO}. The general paradigm of pre-training and fine-tuning has also been proven effective on specific tasks and domains~\cite{Wen2019NeurIPS2R,zhou2021topicbert}.




\subsection{The Fragility of Language Model Fine-tuning}
The phenomenon of instability of PLMs fine-tuning was first reported by Devlin et al.~\cite{Devlin2019BERTPO}. 
A further study~\cite{Dodge2020FineTuningPL} reveals the sensitivity of BERT fine-tuning to random seeds where the randomness is introduced by the shuffle of data order and the random initialization of the task-specific layer by conducting extensive empirical analysis. Inspired by the above experimental analysis, several new methods have been proposed to mitigate the fragility of language model fine-tuning. Mixout \cite{Lee2020MixoutER} replaces parameters with their pre-trained value with a probability p in the fine-tuning phase to promote both the stability and performance of the BERT model. Zhang et al.~\cite{Zhang2020RevisitingFB} figure out that  debiasing the Adam optimizer is beneficial for BERT fine-tuning \cite{Kingma2015AdamAM} through experiments and point out that re-initialize some top layers of a BERT model contributes to better generalization to the downstream tasks. Mosbach et al.~\cite{Mosbach2020OnTS} discuss the reason for the instability of fine-tuning via extensive experiments and suggest using a small learning rate as well as bias correction to improve the generalizability of language model fine-tuning.


\subsection{Regularization}
Regularization is a widely adopted method for improving the performance of deep neural networks. In transfer learning, the most common problems are overfitting and catastrophic forgetting, while a regularization item can help mitigate these issues. Several regularization methods have been proposed to improve the performance of PLMs' fine-tuning. Loshchilov and  Hutter~\cite{Loshchilov2019DecoupledWD} propose a decoupled weight decay regularizer integrated with Adam \cite{Kingma2015AdamAM} optimizer to prevent neural networks from being too complicated. In addition, spectral-norm-based regularization methods~\cite{Yoshida2017SpectralNR, Roth2019AdversarialTG} can be regarded as a general method to constrain the Lipschitz continuity of neural networks that can help to improve the smoothness of the learned neural networks. In addition, \cite{ro2021rollback} proposed to roll back pre-trained weights as an implicit form of regularization to pursue flatter local minima for fine-tuning.

Recently, several approaches relevant to input noise have emerged to improve the local Lipschitz continuity of language models and hence improve the smoothness and generalizability of the fine-tuned model. These algorithms usually minimize the maximum risk caused by noise perturbations within a norm ball. Such approaches include SMART \cite{Jiang2020SMARTRA}, FreeLB \cite{Zhu2020FreeLBEA} and R3F \cite{Aghajanyan2020BetterFB}. They achieve state-of-the-art performance on GLUE, SciTail \cite{Khot2018SciTaiLAT}, and LAMA \cite{petroni2019language,petroni2020context}, etc. NLU benchmarks. 


\section{Conclusion}
In this paper, we investigate the problem of fine-tuning pre-trained language models from the perspective of noise stability. We introduce a lightweight and effective framework, named Layerwise Noise Stability Regularization (LNSR), to improve generalizability and stability when fine-tuning pre-trained language models on a few training samples. 
In our proposed LNSR framework, two alternative noise sampling strategies are used, which are the standard Gaussian noise and In-manifold noise. Our proposed LNSR methods are general techniques that promote the smoothness of language models and thus improve the model's performance. Furthermore, we theoretically analyze the properties of our proposed model connected to the Lipschitz continuity and Tikhonov regularizer. In addition, the experimental results in this paper also reflect the effectiveness of our proposed method to improve the generalizability and stability of pre-trained language models.
\ifCLASSOPTIONcompsoc
  \section*{Acknowledgments}
\else
  \section*{Acknowledgment}
\fi

We would like to thank the Jeffries Data Science Fellowship for supporting Hang Hua's research.

\bibliographystyle{IEEEtran}
\bibliography{TKDE}

\begin{thebibliography}{10}
\providecommand{\url}[1]{#1}
\csname url@samestyle\endcsname
\providecommand{\newblock}{\relax}
\providecommand{\bibinfo}[2]{#2}
\providecommand{\BIBentrySTDinterwordspacing}{\spaceskip=0pt\relax}
\providecommand{\BIBentryALTinterwordstretchfactor}{4}
\providecommand{\BIBentryALTinterwordspacing}{\spaceskip=\fontdimen2\font plus
\BIBentryALTinterwordstretchfactor\fontdimen3\font minus \fontdimen4\font\relax}
\providecommand{\BIBforeignlanguage}[2]{{%
\expandafter\ifx\csname l@#1\endcsname\relax
\typeout{** WARNING: IEEEtran.bst: No hyphenation pattern has been}%
\typeout{** loaded for the language `#1'. Using the pattern for}%
\typeout{** the default language instead.}%
\else
\language=\csname l@#1\endcsname
\fi
#2}}
\providecommand{\BIBdecl}{\relax}
\BIBdecl

\bibitem{Li2018ExplicitIB}
X.~Li, Y.~Grandvalet, and F.~Davoine, ``Explicit inductive bias for transfer learning with convolutional networks,'' in \emph{ICML}, 2018.

\bibitem{Lee2020MixoutER}
C.~Lee, K.~Cho, and W.~Kang, ``Mixout: Effective regularization to finetune large-scale pretrained language models,'' \emph{ArXiv}, vol. abs/1909.11299.

\bibitem{Zhu2020FreeLBEA}
C.~Zhu, Y.~Cheng, Z.~Gan, S.~Sun, T.~Goldstein, and J.~jing Liu, ``Freelb: Enhanced adversarial training for natural language understanding,'' \emph{arXiv: Computation and Language}, 2020.

\bibitem{Jiang2020SMARTRA}
H.~Jiang, P.~He, W.~Chen, X.~Liu, J.~Gao, and T.~Zhao, ``Smart: Robust and efficient fine-tuning for pre-trained natural language models through principled regularized optimization,'' in \emph{ACL}, 2020.

\bibitem{Guu2020REALMRL}
K.~Guu, K.~Lee, Z.~Tung, P.~Pasupat, and M.-W. Chang, ``Realm: Retrieval-augmented language model pre-training,'' \emph{ArXiv}, vol. abs/2002.08909, 2020.

\bibitem{Liu2019FinetuneBF}
Y.~Liu, ``Fine-tune bert for extractive summarization,'' \emph{ArXiv}, vol. abs/1903.10318, 2019.

\bibitem{Wadden2019EntityRA}
D.~Wadden, U.~Wennberg, Y.~Luan, and H.~Hajishirzi, ``Entity, relation, and event extraction with contextualized span representations,'' in \emph{EMNLP/IJCNLP}, 2019.

\bibitem{Zhu2020IncorporatingBI}
J.~Zhu, Y.~Xia, L.~Wu, D.~He, T.~Qin, W.~Zhou, H.~Li, and T.~Liu, ``Incorporating bert into neural machine translation,'' \emph{ArXiv}, vol. abs/2002.06823, 2020.

\bibitem{clark2020electra}
K.~Clark, M.-T. Luong, Q.~V. Le, and C.~D. Manning, ``Electra: Pre-training text encoders as discriminators rather than generators,'' \emph{arXiv preprint arXiv:2003.10555}, 2020.

\bibitem{joshi2020spanbert}
M.~Joshi, D.~Chen, Y.~Liu, D.~S. Weld, L.~Zettlemoyer, and O.~Levy, ``Spanbert: Improving pre-training by representing and predicting spans,'' \emph{Transactions of the Association for Computational Linguistics}, 2020.

\bibitem{2020t5}
\BIBentryALTinterwordspacing
C.~Raffel, N.~Shazeer, A.~Roberts, K.~Lee, S.~Narang, M.~Matena, Y.~Zhou, W.~Li, and P.~J. Liu, ``Exploring the limits of transfer learning with a unified text-to-text transformer,'' \emph{Journal of Machine Learning Research}, vol.~21, no. 140, pp. 1--67, 2020. [Online]. Available: \url{http://jmlr.org/papers/v21/20-074.html}
\BIBentrySTDinterwordspacing

\bibitem{lewis2019bart}
M.~Lewis, Y.~Liu, N.~Goyal, M.~Ghazvininejad, A.~Mohamed, O.~Levy, V.~Stoyanov, and L.~Zettlemoyer, ``Bart: Denoising sequence-to-sequence pre-training for natural language generation, translation, and comprehension,'' \emph{arXiv preprint arXiv:1910.13461}, 2019.

\bibitem{Wang2018GLUEAM}
A.~Wang, A.~Singh, J.~Michael, F.~Hill, O.~Levy, and S.~R. Bowman, ``Glue: A multi-task benchmark and analysis platform for natural language understanding,'' \emph{ArXiv}, vol. abs/1804.07461, 2018.

\bibitem{Wang2019SuperGLUEAS}
A.~Wang, Y.~Pruksachatkun, N.~Nangia, A.~Singh, J.~Michael, F.~Hill, O.~Levy, and S.~R. Bowman, ``Superglue: A stickier benchmark for general-purpose language understanding systems,'' \emph{ArXiv}, vol. abs/1905.00537, 2019.

\bibitem{petroni2019language}
F.~Petroni, T.~Rockt{\"a}schel, P.~Lewis, A.~Bakhtin, Y.~Wu, A.~H. Miller, and S.~Riedel, ``Language models as knowledge bases?'' \emph{arXiv preprint arXiv:1909.01066}, 2019.

\bibitem{petroni2020context}
F.~Petroni, P.~Lewis, A.~Piktus, T.~Rockt{\"a}schel, Y.~Wu, A.~H. Miller, and S.~Riedel, ``How context affects language models' factual predictions,'' \emph{arXiv preprint arXiv:2005.04611}, 2020.

\bibitem{lample2019cross}
G.~Lample and A.~Conneau, ``Cross-lingual language model pretraining,'' \emph{arXiv preprint arXiv:1901.07291}, 2019.

\bibitem{hu2022promptcap}
Y.~Hu, H.~Hua, Z.~Yang, W.~Shi, N.~A. Smith, and J.~Luo, ``Promptcap: Prompt-guided task-aware image captioning,'' \emph{arXiv preprint arXiv:2211.09699}, 2022.

\bibitem{lin2023videoxum}
J.~Lin, H.~Hua, M.~Chen, Y.~Li, J.~Hsiao, C.~Ho, and J.~Luo, ``Videoxum: Cross-modal visual and textural summarization of videos,'' \emph{arXiv preprint arXiv:2303.12060}, 2023.

\bibitem{zhang2020graph}
J.~Zhang, H.~Zhang, C.~Xia, and L.~Sun, ``Graph-bert: Only attention is needed for learning graph representations,'' \emph{arXiv preprint arXiv:2001.05140}, 2020.

\bibitem{Devlin2019BERTPO}
J.~Devlin, M.-W. Chang, K.~Lee, and K.~Toutanova, ``Bert: Pre-training of deep bidirectional transformers for language understanding,'' in \emph{NAACL-HLT}, 2019.

\bibitem{Dodge2020FineTuningPL}
J.~Dodge, G.~Ilharco, R.~Schwartz, A.~Farhadi, H.~Hajishirzi, and N.~A. Smith, ``Fine-tuning pretrained language models: Weight initializations, data orders, and early stopping,'' \emph{ArXiv}, vol. abs/2002.06305, 2020.

\bibitem{bishop1995training}
C.~M. Bishop, ``Training with noise is equivalent to tikhonov regularization,'' \emph{Neural computation}, vol.~7, no.~1, pp. 108--116, 1995.

\bibitem{rifai2011adding}
S.~Rifai, X.~Glorot, Y.~Bengio, and P.~Vincent, ``Adding noise to the input of a model trained with a regularized objective,'' \emph{arXiv preprint arXiv:1104.3250}, 2011.

\bibitem{arora2018stronger}
S.~Arora, R.~Ge, B.~Neyshabur, and Y.~Zhang, ``Stronger generalization bounds for deep nets via a compression approach,'' in \emph{International Conference on Machine Learning}.\hskip 1em plus 0.5em minus 0.4em\relax PMLR, 2018, pp. 254--263.

\bibitem{dong2021should}
X.~Dong, A.~T. Luu, M.~Lin, S.~Yan, and H.~Zhang, ``How should pre-trained language models be fine-tuned towards adversarial robustness?'' \emph{Advances in Neural Information Processing Systems}, vol.~34, pp. 4356--4369, 2021.

\bibitem{hua2021noise}
H.~Hua, X.~Li, D.~Dou, C.-Z. Xu, and J.~Luo, ``Noise stability regularization for improving bert fine-tuning,'' \emph{arXiv preprint arXiv:2107.04835}, 2021.

\bibitem{li2022method}
X.~Li, H.~Hang, C.~Xu, and D.~Dou, ``Method and apparatus for transfer learning,'' Dec.~15 2022, uS Patent App. 17/820,321.

\bibitem{federer1996geometric}
H.~Federer \emph{et~al.}, ``Geometric measure theory,'' 1996.

\bibitem{sietsma1991creating}
J.~Sietsma and R.~J. Dow, ``Creating artificial neural networks that generalize,'' \emph{Neural networks}, vol.~4, no.~1, pp. 67--79, 1991.

\bibitem{tikhonov1977solutions}
A.~N. Tikhonov and V.~Y. Arsenin, ``Solutions of ill-posed problems,'' \emph{New York}, vol.~1, no.~30, p. 487, 1977.

\bibitem{roweis2000nonlinear}
S.~T. Roweis and L.~K. Saul, ``Nonlinear dimensionality reduction by locally linear embedding,'' \emph{science}, vol. 290, no. 5500, pp. 2323--2326, 2000.

\bibitem{lin2008riemannian}
T.~Lin and H.~Zha, ``Riemannian manifold learning,'' \emph{IEEE Transactions on Pattern Analysis and Machine Intelligence}, vol.~30, no.~5, pp. 796--809, 2008.

\bibitem{huo2007survey}
X.~Huo, X.~S. Ni, and A.~K. Smith, ``A survey of manifold-based learning methods,'' \emph{Recent advances in data mining of enterprise data}, pp. 691--745, 2007.

\bibitem{cayton2005algorithms}
L.~Cayton, ``Algorithms for manifold learning,'' \emph{Univ. of California at San Diego Tech. Rep}, vol.~12, no. 1-17, p.~1, 2005.

\bibitem{van2008visualizing}
L.~Van~der Maaten and G.~Hinton, ``Visualizing data using t-sne.'' \emph{Journal of machine learning research}, vol.~9, no.~11, 2008.

\bibitem{van2017l2}
T.~Van~Laarhoven, ``L2 regularization versus batch and weight normalization,'' \emph{arXiv preprint arXiv:1706.05350}, 2017.

\bibitem{zhang2018three}
G.~Zhang, C.~Wang, B.~Xu, and R.~Grosse, ``Three mechanisms of weight decay regularization,'' in \emph{International Conference on Learning Representations}, 2018.

\bibitem{szegedy2014intriguing}
C.~Szegedy, W.~Zaremba, I.~Sutskever, J.~Bruna, D.~Erhan, I.~Goodfellow, and R.~Fergus, ``Intriguing properties of neural networks,'' in \emph{2nd International Conference on Learning Representations, ICLR 2014}, 2014.

\bibitem{madry2018towards}
A.~Madry, A.~Makelov, L.~Schmidt, D.~Tsipras, and A.~Vladu, ``Towards deep learning models resistant to adversarial attacks,'' in \emph{International Conference on Learning Representations}, 2018.

\bibitem{Warstadt2019NeuralNA}
A.~Warstadt, A.~Singh, and S.~R. Bowman, ``Neural network acceptability judgments,'' \emph{Transactions of the Association for Computational Linguistics}, vol.~7, pp. 625--641, 2019.

\bibitem{matthews1975comparison}
B.~W. Matthews, ``Comparison of the predicted and observed secondary structure of t4 phage lysozyme,'' \emph{Biochimica et Biophysica Acta (BBA)-Protein Structure}, vol. 405, no.~2, pp. 442--451, 1975.

\bibitem{Dolan2005AutomaticallyCA}
W.~Dolan and C.~Brockett, ``Automatically constructing a corpus of sentential paraphrases,'' in \emph{IWP@IJCNLP}, 2005.

\bibitem{Dagan2005ThePR}
I.~Dagan, O.~Glickman, and B.~Magnini, ``The pascal recognising textual entailment challenge,'' in \emph{MLCW}, 2005.

\bibitem{BarHaim2006TheSP}
R.~Bar-Haim, I.~Dagan, B.~Dolan, L.~Ferro, D.~Giampiccolo, and B.~Magnini, ``The second pascal recognising textual entailment challenge.''

\bibitem{Giampiccolo2007TheTP}
D.~Giampiccolo, B.~Magnini, I.~Dagan, and W.~Dolan, ``The third pascal recognizing textual entailment challenge,'' in \emph{ACL-PASCAL@ACL}, 2007.

\bibitem{Cer2017SemEval2017T1}
D.~M. Cer, M.~T. Diab, E.~Agirre, I.~Lopez-Gazpio, and L.~Specia, ``Semeval-2017 task 1: Semantic textual similarity multilingual and crosslingual focused evaluation,'' \emph{ArXiv}, vol. abs/1708.00055, 2017.

\bibitem{rajpurkar2016squad}
P.~Rajpurkar, J.~Zhang, K.~Lopyrev, and P.~Liang, ``Squad: 100,000+ questions for machine comprehension of text,'' \emph{arXiv preprint arXiv:1606.05250}, 2016.

\bibitem{fisch2019mrqa}
A.~Fisch, A.~Talmor, R.~Jia, M.~Seo, E.~Choi, and D.~Chen, ``Mrqa 2019 shared task: Evaluating generalization in reading comprehension,'' \emph{arXiv preprint arXiv:1910.09753}, 2019.

\bibitem{Liu2019RoBERTaAR}
Y.~Liu, M.~Ott, N.~Goyal, J.~Du, M.~Joshi, D.~Chen, O.~Levy, M.~Lewis, L.~Zettlemoyer, and V.~Stoyanov, ``Roberta: A robustly optimized bert pretraining approach,'' \emph{ArXiv}, vol. abs/1907.11692, 2019.

\bibitem{Srivastava2014DropoutAS}
N.~Srivastava, G.~E. Hinton, A.~Krizhevsky, I.~Sutskever, and R.~Salakhutdinov, ``Dropout: a simple way to prevent neural networks from overfitting,'' \emph{J. Mach. Learn. Res.}, vol.~15, pp. 1929--1958, 2014.

\bibitem{Wan2013RegularizationON}
L.~Wan, M.~D. Zeiler, S.~Zhang, Y.~LeCun, and R.~Fergus, ``Regularization of neural networks using dropconnect,'' in \emph{ICML}, 2013.

\bibitem{yang2022improving}
C.~Yang and X.~Ma, ``Improving stability of fine-tuning pretrained language models via component-wise gradient norm clipping,'' \emph{arXiv preprint arXiv:2210.10325}, 2022.

\bibitem{Wolf2019HuggingFacesTS}
T.~Wolf, L.~Debut, V.~Sanh, J.~Chaumond, C.~Delangue, A.~Moi, P.~Cistac, T.~Rault, R.~Louf, M.~Funtowicz, J.~Davison, S.~Shleifer, P.~von Platen, C.~Ma, Y.~Jernite, J.~Plu, C.~Xu, T.~L. Scao, S.~Gugger, M.~Drame, Q.~Lhoest, and A.~M. Rush, ``Huggingface's transformers: State-of-the-art natural language processing,'' \emph{ArXiv}, vol. abs/1910.03771, 2019.

\bibitem{Kingma2015AdamAM}
D.~P. Kingma and J.~Ba, ``Adam: A method for stochastic optimization,'' \emph{CoRR}, vol. abs/1412.6980, 2015.

\bibitem{JDH17}
J.~Johnson, M.~Douze, and H.~J{\'e}gou, ``Billion-scale similarity search with gpus,'' \emph{arXiv preprint arXiv:1702.08734}, 2017.

\bibitem{Aghajanyan2020BetterFB}
A.~Aghajanyan, A.~Shrivastava, A.~Gupta, N.~Goyal, L.~Zettlemoyer, and S.~Gupta, ``Better fine-tuning by reducing representational collapse,'' \emph{ArXiv}, vol. abs/2008.03156, 2020.

\bibitem{tsiprasrobustness}
D.~Tsipras, S.~Santurkar, L.~Engstrom, A.~Turner, and A.~Madry, ``Robustness may be at odds with accuracy,'' in \emph{International Conference on Learning Representations}, 2019.

\bibitem{cai2020isotropy}
X.~Cai, J.~Huang, Y.~Bian, and K.~Church, ``Isotropy in the contextual embedding space: Clusters and manifolds,'' in \emph{International Conference on Learning Representations}, 2020.

\bibitem{Erhan2009TheDO}
D.~Erhan, P.-A. Manzagol, Y.~Bengio, S.~Bengio, and P.~Vincent, ``The difficulty of training deep architectures and the effect of unsupervised pre-training,'' in \emph{AISTATS}, 2009.

\bibitem{Erhan2010WhyDU}
D.~Erhan, A.~C. Courville, Y.~Bengio, and P.~Vincent, ``Why does unsupervised pre-training help deep learning?'' \emph{J. Mach. Learn. Res.}, vol.~11, pp. 625--660, 2010.

\bibitem{Pennington2014GloveGV}
J.~Pennington, R.~Socher, and C.~D. Manning, ``Glove: Global vectors for word representation,'' in \emph{EMNLP}, 2014.

\bibitem{Mikolov2013DistributedRO}
T.~Mikolov, I.~Sutskever, K.~Chen, G.~S. Corrado, and J.~Dean, ``Distributed representations of words and phrases and their compositionality,'' \emph{ArXiv}, vol. abs/1310.4546, 2013.

\bibitem{Dai2015SemisupervisedSL}
A.~M. Dai and Q.~V. Le, ``Semi-supervised sequence learning,'' in \emph{NIPS}, 2015.

\bibitem{Peters2018DeepCW}
M.~E. Peters, M.~Neumann, M.~Iyyer, M.~Gardner, C.~Clark, K.~Lee, and L.~Zettlemoyer, ``Deep contextualized word representations,'' \emph{ArXiv}, vol. abs/1802.05365, 2018.

\bibitem{Radford2018ImprovingLU}
A.~Radford, ``Improving language understanding by generative pre-training,'' 2018.

\bibitem{Radford2019LanguageMA}
A.~Radford, J.~Wu, R.~Child, D.~Luan, D.~Amodei, and I.~Sutskever, ``Language models are unsupervised multitask learners,'' 2019.

\bibitem{brown2020language}
T.~Brown, B.~Mann, N.~Ryder, M.~Subbiah, J.~D. Kaplan, P.~Dhariwal, A.~Neelakantan, P.~Shyam, G.~Sastry, A.~Askell \emph{et~al.}, ``Language models are few-shot learners,'' \emph{Advances in neural information processing systems}, vol.~33, pp. 1877--1901, 2020.

\bibitem{Yang2019XLNetGA}
Z.~Yang, Z.~Dai, Y.~Yang, J.~Carbonell, R.~Salakhutdinov, and Q.~V. Le, ``Xlnet: Generalized autoregressive pretraining for language understanding,'' in \emph{NeurIPS}, 2019.

\bibitem{Wen2019NeurIPS2R}
\BIBentryALTinterwordspacing
Z.~Wen, S.-C. Fuh, and A.~Mircea, ``Neurips 2019 reproducibility challenge: Controllable unsupervised text attribute transfer via editing entangled latent representation,'' 2019. [Online]. Available: \url{https://api.semanticscholar.org/CorpusID:211094926}
\BIBentrySTDinterwordspacing

\bibitem{zhou2021topicbert}
Y.~Zhou, L.~Liao, Y.~Gao, R.~Wang, and H.~Huang, ``Topicbert: A topic-enhanced neural language model fine-tuned for sentiment classification,'' \emph{IEEE Transactions on Neural Networks and Learning Systems}, 2021.

\bibitem{Zhang2020RevisitingFB}
T.~Zhang, F.~Wu, A.~Katiyar, K.~Q. Weinberger, and Y.~Artzi, ``Revisiting few-sample bert fine-tuning,'' \emph{ArXiv}, vol. abs/2006.05987, 2020.

\bibitem{Mosbach2020OnTS}
M.~Mosbach, M.~Andriushchenko, and D.~Klakow, ``On the stability of fine-tuning bert: Misconceptions, explanations, and strong baselines,'' \emph{ArXiv}, vol. abs/2006.04884, 2020.

\bibitem{Loshchilov2019DecoupledWD}
I.~Loshchilov and F.~Hutter, ``Decoupled weight decay regularization,'' in \emph{ICLR}, 2019.

\bibitem{Yoshida2017SpectralNR}
Y.~Yoshida and T.~Miyato, ``Spectral norm regularization for improving the generalizability of deep learning,'' \emph{ArXiv}, vol. abs/1705.10941, 2017.

\bibitem{Roth2019AdversarialTG}
K.~Roth, Y.~Kilcher, and T.~Hofmann, ``Adversarial training generalizes data-dependent spectral norm regularization,'' \emph{ArXiv}, vol. abs/1906.01527, 2019.

\bibitem{ro2021rollback}
Y.~Ro, J.~Choi, B.~Heo, and J.~Y. Choi, ``Rollback ensemble with multiple local minima in fine-tuning deep learning networks,'' \emph{IEEE transactions on neural networks and learning systems}, vol.~33, no.~9, pp. 4648--4660, 2021.

\bibitem{Khot2018SciTaiLAT}
T.~Khot, A.~Sabharwal, and P.~Clark, ``Scitail: A textual entailment dataset from science question answering,'' in \emph{AAAI}, 2018.

\bibitem{tsatsaronis2015overview}
G.~Tsatsaronis, G.~Balikas, P.~Malakasiotis, I.~Partalas, M.~Zschunke, M.~R. Alvers, D.~Weissenborn, A.~Krithara, S.~Petridis, D.~Polychronopoulos \emph{et~al.}, ``An overview of the bioasq large-scale biomedical semantic indexing and question answering competition,'' \emph{BMC bioinformatics}, vol.~16, no.~1, pp. 1--28, 2015.

\bibitem{dua2019drop}
D.~Dua, Y.~Wang, P.~Dasigi, G.~Stanovsky, S.~Singh, and M.~Gardner, ``Drop: A reading comprehension benchmark requiring discrete reasoning over paragraphs,'' \emph{arXiv preprint arXiv:1903.00161}, 2019.

\bibitem{saha2018duorc}
A.~Saha, R.~Aralikatte, M.~M. Khapra, and K.~Sankaranarayanan, ``Duorc: Towards complex language understanding with paraphrased reading comprehension,'' \emph{arXiv preprint arXiv:1804.07927}, 2018.

\bibitem{lai2017race}
G.~Lai, Q.~Xie, H.~Liu, Y.~Yang, and E.~Hovy, ``Race: Large-scale reading comprehension dataset from examinations,'' \emph{arXiv preprint arXiv:1704.04683}, 2017.

\bibitem{levy2017zero}
O.~Levy, M.~Seo, E.~Choi, and L.~Zettlemoyer, ``Zero-shot relation extraction via reading comprehension,'' \emph{arXiv preprint arXiv:1706.04115}, 2017.

\bibitem{kembhavi2017you}
A.~Kembhavi, M.~Seo, D.~Schwenk, J.~Choi, A.~Farhadi, and H.~Hajishirzi, ``Are you smarter than a sixth grader? textbook question answering for multimodal machine comprehension,'' in \emph{Proceedings of the IEEE Conference on Computer Vision and Pattern recognition}, 2017.

\end{thebibliography}






\clearpage
\appendices

\section{Noise Stability of Deep CNNs}
\label{sec:noise_cnn}
\begin{figure}[h!]
  \centering
  \includegraphics[width=0.85\linewidth]{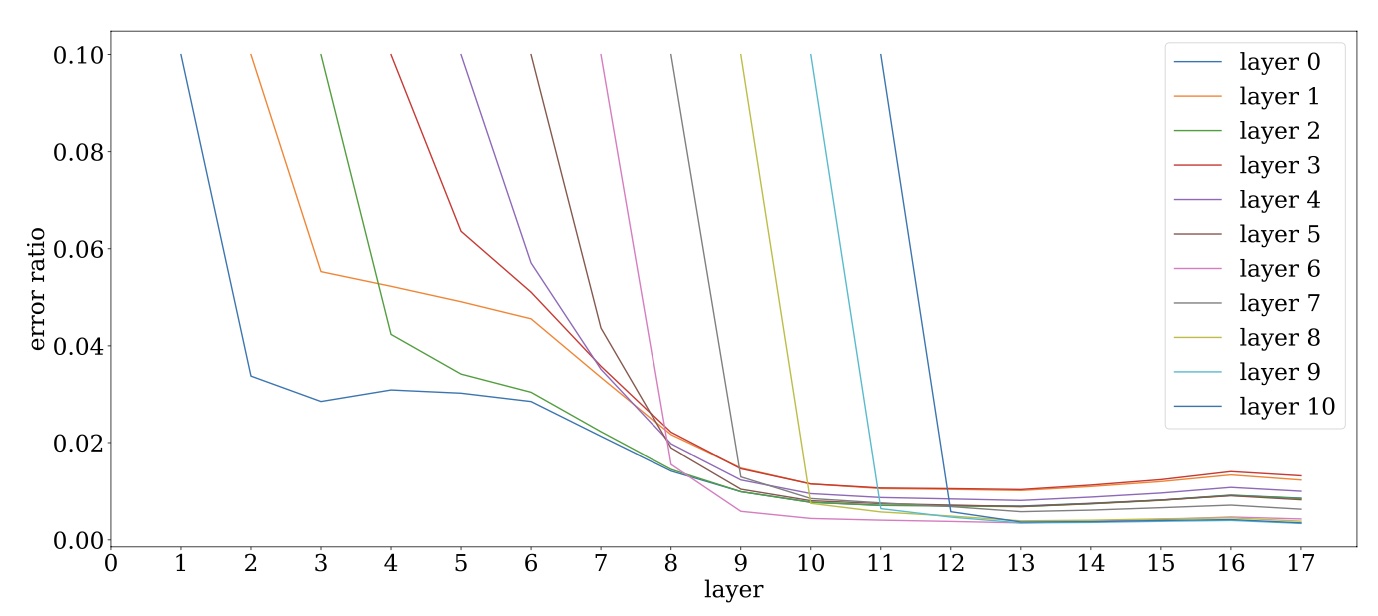}
  \caption{Attenuation of injected noise on a VGG-19 net trained on CIFAR-10. The x-axis is the index of layers and the y-axis denotes the relative error introduced by the noise  ($\|\hat{x}^i-x^i\|_2/\|x^i\|_2$). A curve starts at the layer where a scaled Gaussian noise is injected into its input, whose l2 norm is set to 10\% of the norm of its original input. As it propagates up, the injected noise has a rapidly decreasing effect on higher layers.}
  \label{fig:noise_vgg19}
\end{figure}

Here we demonstrate that deep CNNs are much more resilient to injected noise than transformer-based architectures such as BERT. An example on VGG-19 is shown in Figure~\ref{fig:noise_vgg19}. This figure comes from \cite{arora2018stronger}.

\vspace{-0.3cm}

\section{Introduction for the Experimental Datasets}
\label{sec:experimental}
\subsection{The Selected Text Classification Tasks}
\begin{center}
\begin{table}[h!]
\resizebox{\linewidth}{!}{
  \begin{tabular}{lllllll}
    \toprule
    \multirow{2}{*}{} && RTE & MRPC & CoLA & STS-B \\
\midrule
    &Task& NLI & Paraphrase & Acceptability & Similarity \\
    & Metrics & Accuracy & Accuracy/F1 & Matthews Corr & $\frac{\text{Pearson+Spearman corr}}{2}$\\
    & $\#$ of labels & 2 & 2 & 2 & 1\\
    & $\#$ of training samples & 2.5k & 3.7k & 8.6k & 7k\\
    & $\#$ of validation samples & 276 & 408 & 1k & 1.5k\\
    & $\#$ of test samples & 3k & 1.7k & 1k & 1.4k\\
    \bottomrule
  \end{tabular}
  }
    \caption{The summarization of the selected GLUE benchmark tasks used in this work.}
    \label{tab:datasets}
\end{table}
\end{center}
\vspace{-1.2cm}

\subsection{Out-of-domain Question Answering Tasks}
\label{sec:experimental}
\textbf{BioASQ} \cite{tsatsaronis2015overview} is a
large-scale semantic indexing and question answering dataset in the biomedical domain, all the question, and answer pairs are created by domain experts. 

\textbf{DROP} \cite{dua2019drop} examples are collected similarly to SQuAD, where the question-answer pairs are constructed from Wikipedia paragraphs by crowd workers. Different from SQuAD, the questions of DROP mainly focus on quantitative reasoning. Besides, DROP contains non-extractive numeric answers as well as extractive text answers.

\textbf{DuoRC} \cite{saha2018duorc} is a paraphrase-based reading comprehension dataset. It contains 186089 QA pairs created from a collection of paraphrased movie plots. The main challenge of this task is that it requires models to go beyond the content of the given passage and incorporate different kinds of knowledge to arrive at the answer.

\textbf{RACE} \cite{lai2017race} is a machine reading comprehension dataset that is collected from English reading comprehension exams for middle and high school students in China.

\textbf{RelationExtraction} \cite{levy2017zero} is a multi-turn question answering task which is built from relation extraction datasets (e.g. ACE04, ACE05, and CoNLL04). The entities and relations in the relation extraction datasets are transformed into question-answer pairs using templates.

\textbf{TextbookQA} \cite{kembhavi2017you} is a QA dataset drawn from middle school science curricula. It contains 12567 questions in total. 

\begin{figure*}[t!]
  \centering
  \includegraphics[width=0.7\linewidth]{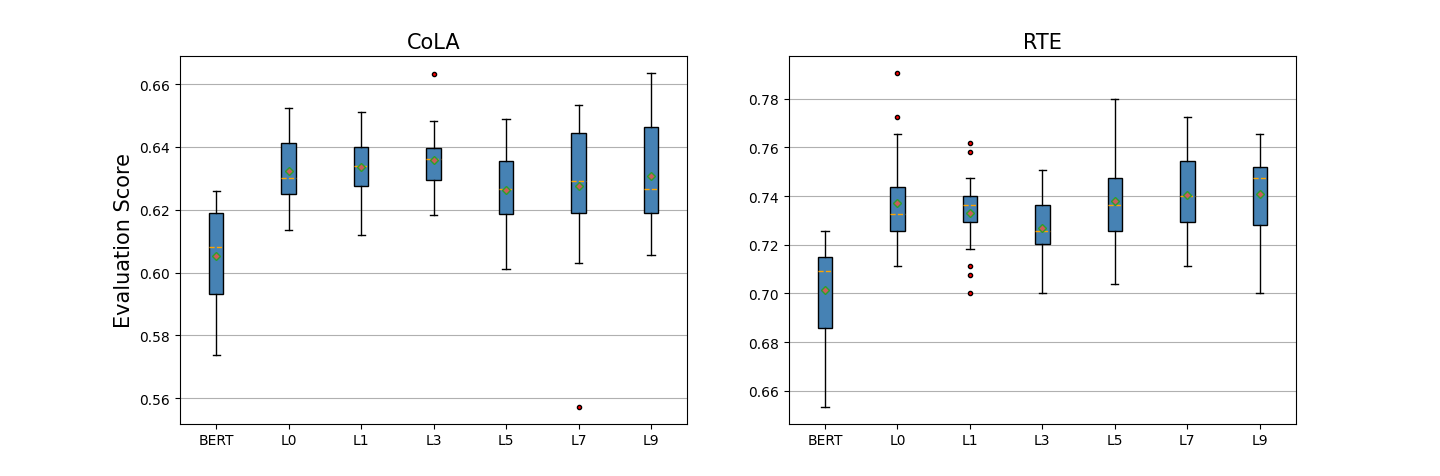}
  \includegraphics[width=0.7\linewidth]{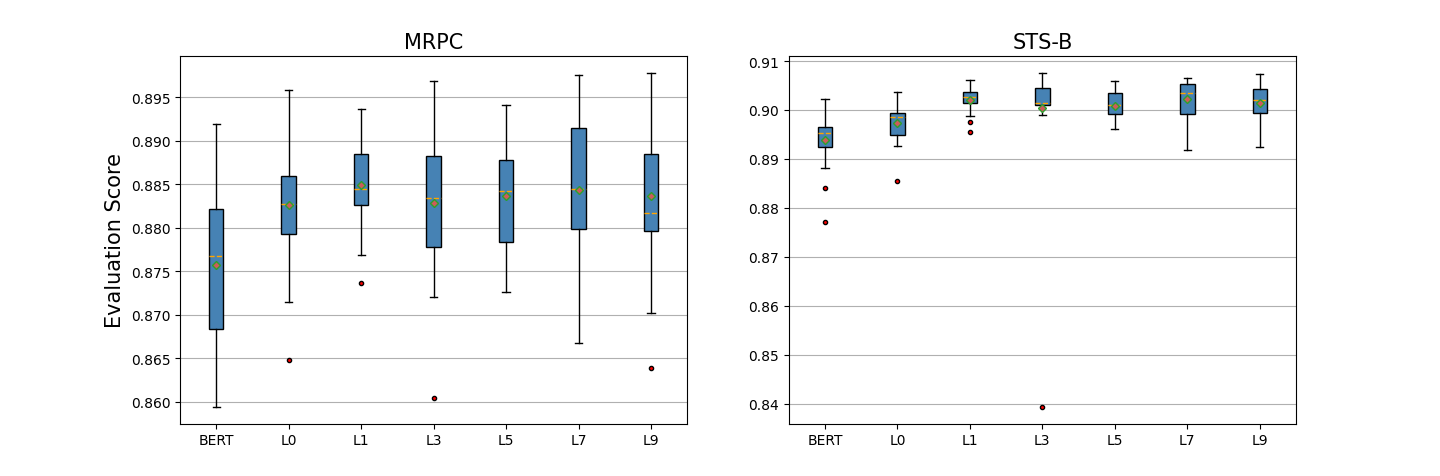}
  \caption{Performance distribution of the BERT model with different noise injection positions across 25 random seeds.}
  \label{fig:sens}
\end{figure*}

 \begin{center}
\begin{table*}[t!]
  \label{tab:commands}
  \resizebox{\textwidth}{!}{
  \begin{tabular}{llllllllllllll}
    \toprule

\multirow{2}{*}{}& \multicolumn{3}{c}{RTE} & \multicolumn{3}{c}{MRPC} & \multicolumn{3}{c}{CoLA}& \multicolumn{3}{c}{STS-B} \\
\cmidrule(r){2-4} \cmidrule(r){5-7} \cmidrule(r){8-10} \cmidrule(r){11-13}
\makecell[l]{$\text{BERT}_{\text{LARGE}}$} &  train      &  dev   &   gap $\downarrow$
&  train      &  dev   &   gap $\downarrow$
&  train      &  dev   &   gap $\downarrow$
&  train      &  dev   &   gap $\downarrow$\\
\midrule
    FT \cite{Devlin2019BERTPO}& $\bf{95.89}$&$70.13$&$25.76$& $96.57$&$87.57$&$9.00$&
    $\bf{97.71}$&$61.56 $&$36.25$ &
    $98.31 $&$89.38 $& $8.93$ \\
    \midrule
    \bf{Standard LNSR} & 
    $90.72$&$73.31$&$17.41$& $96.68$&$88.50$&$\bf{8.18} $&
    $93.44 $&$63.35 $&$\bf{30.09}$ &
    $\bf{98.45} $&$90.23 $& $8.22$\\
    \textbf{In-manifold LNSR}&
    $89.52$&$\bf{75.09}$&$\bf{14.43}$& $\bf98.42$&$\bf{88.85}$&$9.57 $&
    $96.01 $&$\bf{63.64}$&$32.65$ &
    $96.77 $&$\bf{91.05}$& $\bf{5.75}$\\
    \bottomrule
  \end{tabular}
  }
    \caption{Comparison of the generalizability performance of different models. We report the mean training/evaluation Acc and the generalizability gap (training Acc - evaluation Acc) of each model across 20 random seeds.}
    \label{tab:geng}
\end{table*}
\end{center}

\vspace{-1cm}
\section{Other Experimental Analysis}
\label{sec:appc}
In this section, we show the effects of the position of Gaussian noise injection on models' performance. The interpretation of Figure~\ref{fig:sens} is described in section 5.3. In addition, Table~\ref{tab:geng} shows the mean training/evaluation accuracy and generalization gap of different methods on each task. We can conclude that fine-tuning with LNSR and In-manifold LNSR can effectively narrow the generalization gap and help improve the performance.

\end{document}